\documentclass[sn-mathphys-num]{sn-jnl}% Math and Physical Sciences Numbered Reference Style 
\usepackage{graphicx}%
\usepackage{multirow}%
\usepackage{amsmath,amssymb,amsfonts}%
\usepackage{amsthm}%
\usepackage{mathrsfs}%
\usepackage[title]{appendix}%
\usepackage{xcolor}%
\usepackage{textcomp}%
\usepackage{manyfoot}%
\usepackage{booktabs}%
\usepackage{algorithm}%
\usepackage{algorithmic}%
\usepackage{listings}%
\theoremstyle{thmstyleone}%
\newtheorem{theorem}{Theorem}%  meant for continuous numbers
\theoremstyle{thmstyletwo}%
\newtheorem{remark}{Remark}%

\theoremstyle{thmstylethree}%

\newtheorem{lemma}{Lemma}
\newtheorem{corollary}{Corollary}

\begin{document}

\title{FigBO: A Generalized Acquisition Function Framework with Look-Ahead Capability for Bayesian Optimization}


\author[1]{\fnm{Hui} \sur{Chen}}\email{hui.chen2@students.mq.edu.au}

\author[1]{\fnm{Xuhui} \sur{Fan}}\email{xuhui.fan@mq.edu.au}

\author[1]{\fnm{Zhangkai} \sur{Wu}}\email{zhangkai.wu@mq.edu.au}

\author*[1]{\fnm{Longbing} \sur{Cao}}\email{longbing.cao@mq.edu.au}

\affil*[1]{\orgdiv{School of Computing}, \orgname{Macquarie University}, \orgaddress{\city{Sydney}, \postcode{2109}, \country{Australia}}}

% \affil[2]{\orgdiv{Department}, \orgname{Organization}, \orgaddress{\street{Street}, \city{City}, \postcode{10587}, \state{State}, \country{Country}}}

% \affil[3]{\orgdiv{Department}, \orgname{Organization}, \orgaddress{\street{Street}, \city{City}, \postcode{610101}, \state{State}, \country{Country}}}


\abstract{Bayesian optimization is a powerful technique for optimizing expensive-to-evaluate black-box functions, consisting of two main components: a surrogate model and an acquisition function. In recent years, myopic acquisition functions have been widely adopted for their simplicity and effectiveness. However, their lack of look-ahead capability limits their performance. To address this limitation, we propose FigBO, a generalized acquisition function that incorporates the future impact of candidate points on global information gain. FigBO is a plug-and-play method that can integrate seamlessly with most existing myopic acquisition functions. Theoretically, we analyze the regret bound and convergence rate of FigBO when combined with the myopic base acquisition function expected improvement (EI), comparing them to those of standard EI. Empirically, extensive experimental results across diverse tasks demonstrate that FigBO achieves state-of-the-art performance and significantly faster convergence compared to existing methods.}




\maketitle

\section{Introduction}\label{sec1}
Bayesian optimization (BO) \cite{garnett2023bayesian,hayashi2022bayesian,candelieri2024fair} is an effective tool to optimize expensive-to-evaluate black-box functions and has successfully been applied to various problems, including machine learning hyperparameter optimization \cite{bergstra2011algorithms,hvarfner2022pi,vsehic2022lassobench}, neural architecture search \cite{ru2020interpretable,wang2021sample}, and robotics \cite{berkenkamp2016bayesian,calandra2014bayesian}. BO identifies the optimal solution through an iterative process consisting of two key steps: fitting a surrogate model, typically a Gaussian process (GP), to approximate the objective function, and optimizing an acquisition function to guide the selection of the next query point.

The choice of the acquisition function plays a critical role in determining the optimization performance. Depending on whether an acquisition function considers the future impact of candidate points—that is, whether it has a look-ahead capability—it can be categorized into two types: myopic and non-myopic \cite{ziomek2023random}. Compared to non-myopic acquisition functions, myopic ones are computationally efficient and easy to implement while still delivering satisfactory performance, making them the dominant choice in current optimization methods \cite{eriksson2021high,ament2023unexpected,hvarfner2022pi}. 

However, myopic acquisition functions focus solely on short-term optimization gains, i.e., they lack a look-ahead capability, which degrades optimization performance \cite{jones1998efficient,srinivas2012information,thompson1933likelihood}. In contrast, non-myopic acquisition functions take future optimization gains into account when selecting candidate points but have limited applications due to their high computational overhead and highly complex implementation processes \cite{hennig2012entropy,hvarfner2022joint}. Therefore, designing a new acquisition function that balances future optimization benefits with simplicity of implementation has become a significant challenge.

To address this challenge, we propose a new generalized acquisition function framework, FigBO, which equips various \textit{myopic base acquisition functions} with look-ahead capability by incorporating the future impact of candidate points on global information gain. FigBO is a conceptually simple plug-and-play approach that works seamlessly with most myopic acquisition functions while maintaining a balance between exploration and exploitation. 

Formally, we make the following contributions:
\begin{itemize}
    \item We propose a novel generalized version of the acquisition function, FigBO, which grants the myopic base acquisition function look-ahead capability by accounting for the prospective influence of candidate points on global information gain. Moreover, FigBO has been demonstrated to effectively strike a balance between long-term optimization gains and ease of implementation.
    \item We provide the regret bound and convergence rate of FigBO when EI is chosen as the myopic base acquisition function, and demonstrate that the convergence rate of FigBO is asymptotically equivalent to that of the vanilla EI in \cite{bull2011convergence}.
    \item We empirically evaluate the performance of FigBO on different tasks. The experimental results show that FigBO achieves state-of-the-art performance and converges significantly faster than existing methods.
\end{itemize}

\section{Background}
In this section, we introduce the background material for BO. We aim to solve the optimization problem $\mathbf{x}^* \in \mathrm{argmin}_{\mathbf{x}\in \mathcal X} f(\mathbf{x})$ for an expensive-to-evaluate black-box function $f: \mathcal{X} \rightarrow \mathbb{R}$ over the input space $\mathcal{X}\subseteq \mathbb{R}^d$. In our setting, the $f(\mathbf{x})$ can be observed through a noise-corrupted estimate, $y=f(\mathbf{x})+\epsilon$, where $\epsilon \sim \mathcal{N}(0,\sigma^2_{\epsilon})$. BO aims to globally minimize $f$ by starting with an initial design and then sequentially selecting new points $\mathbf{x}_n$ for the iteration $n\in \{1, \dots, N\}$, constructing the data $\mathcal{D}_n=\mathcal{D}_{n-1} \cup \{(\mathbf{x}_n,y_n)\}$ \cite{brochu2010tutorial,hvarfner2023general}. After each new observation, BO employs a probabilistic surrogate model of $f$ \cite{snoek2012practical,hutter2011sequential} and uses this model to build an acquisition function $\alpha(\mathbf{x};\mathcal{D}_n)$, which selects the next query point by balancing exploration and exploitation.   

\subsection{Gaussian Processes}
\textit{Gaussian process} (GP) \cite{williams2006gaussian} is the most commonly used surrogate model. The GP employs a mean function $m(\mathbf{x})$ and covariance kernel $k(\mathbf{x},\mathbf{x'})$ to encode our prior belief on the smoothness of the black-box function $f$. Given the observation $\mathcal{D}_n$ collected so far at iteration $n$, the Gaussian posterior over the objective can be given by $p(f|\mathcal{D}_n)\sim \mathcal{N}(\mu_n(\mathbf{x}),s^2_n(\mathbf{x}))$ with 
\begin{equation}
\mu_n(\mathbf{x})=\mathbf{k}_n(\mathbf{x})^\top (\mathbf{K}_n+\sigma_{\epsilon}^2 \mathbf{I})^{-1} \mathbf{y},
\end{equation}
\begin{equation}
    s_n^2(\mathbf{x})=k(\mathbf{x},\mathbf{x})-\mathbf{k}_n(\mathbf{x})^\top  (\mathbf{K}_n+\sigma^2_{\epsilon} \mathbf{I})^{-1} \mathbf{k}_n(\mathbf{x}),
\end{equation}
where $\mathbf{K}_n$ calculates the covariance kernel on all data pair in $\mathcal{D}_n$, $\mathbf{k}_n(\mathbf{x})$ evaluates the kernel between the $\mathbf{x}$ and all inputs in $\mathcal{D}_n$ \cite{ziomek2023random} and $\sigma^2_{\epsilon}$ denotes the noise variance. In this paper, we use GP as our default surrogate model, and leave other surrogate models including random forest \cite{breiman2001random} and Bayesian neural networks \cite{blundell2015weight} for future work.

\subsection{Acquisition Functions}
Based on the Gaussian posterior above, BO maximizes the acquisition function to find a new query point in an explore-exploit trade-off manner at each iteration. Various myopic and non-myopic acquisition functions are used in BO \cite{garnett2023bayesian}. Among these, the \textit{expected improvement} (EI) \cite{jones1998efficient}, a myopic function, is the most widely used. For a noise-free function, EI chooses the next query point $\mathbf{x}_{n+1}$ as
\begin{equation}
\mathbf{x}_{n+1}\in \mathrm{argmax}_{\mathbf{x}\in \mathcal{X}} \mathbb{E}_{f \sim p(f|\mathcal{D}_n)} [[f(\mathbf{x})-f_n^*]^+]
=\mathrm{argmax}_{\mathbf{x}\in \mathcal{X}}Zs_n(\mathbf{x})\Phi(Z)+s_n(\mathbf{x})\phi(Z), 
\end{equation}
where the $f_n^*$ denotes the best function evaluation observed by iteration $n$ and $Z=(\mu_n(\mathbf{x})-f_n^*)/s_n(\mathbf{x})$. Besides EI, other myopic acquisition functions include upper conﬁdence bound (UCB) \cite{srinivas2012information}, probability of improvement (PI) \cite{jones2001taxonomy} and Thompson sampling (TS) \cite{thompson1933likelihood}, while the non-myopic acquisition functions involve entropy search (ES) \cite{hennig2012entropy}, predictive entropy search (PES) \cite{hernandez2014predictive}, max-value entropy search (MES) \cite{wang2017max}, joint entropy search (JES) \cite{hvarfner2022joint} and knowledge gradient \cite{frazier2008knowledge}. In this paper, we focus on myopic acquisition functions.

\section{Methodology}
In this section, we first explain the crucial role of global information gain in designing acquisition functions and its intrinsic connection to global uncertainty. Then, we introduce our proposed method, FigBO—a simple yet effective plug-and-play framework that seamlessly integrates with most existing myopic acquisition functions.

\subsection{Global Information Gain}
To learn the function $f$ as quickly as possible, it is crucial for sampled points to provide the maximum amount of information, which can be measured by \emph{information gain} \cite{srinivas2012information,hvarfner2024vanilla}. Given a set of observations $\mathbf{y}_{\mathcal{A}}$, the information gain represents the mutual information between the $f$ and observation $\mathbf{y}_{\mathcal{A}}$:
\begin{equation}
I(f;\mathbf{y}_{\mathcal{A}})=H(f)-H(f|\mathbf{y}_{\mathcal{A}}),
\end{equation}
which measures the reduction amount of uncertainty about $f$ due to $\mathbf{y}_{\mathcal{A}}$ being revealed. Assume that $f$ is generated from a GP, we can then rewrite the information gain as  
\begin{equation}
I(f;\mathbf{y}_{A})=\frac{1}{2}\log |I+\sigma_{\epsilon}^{-2}\mathbf{K}|,
\end{equation}
where $\mathbf{K}=k(\mathbf{X}_{\mathcal{A}},\mathbf{X}_{\mathcal{A}})$ denotes the Gram matrix for $\mathbf{X}_{\mathcal{A}}$. Since it is an NP-hard problem \cite{ko1995exact} to obtain the maximizer among ${\mathcal{A}} \subset \mathcal{X}$, this problem is reformulated as a greedy algorithm, which can be written at iteration $n$ by
\begin{equation}
\mathbf{x}_n=\mathrm{argmax}_{\mathbf{x}\in \mathcal{X}} s_{n-1}(\mathbf{x}),
\end{equation}
which means that selecting the point with the highest variance can achieve the maximum information gain by reducing the most uncertainty. However, this is a myopic strategy and only considers the impact of the uncertainty of a single candidate point, i.e. \textit{ local information gain}.

\begin{figure*}[t]
\centering
\includegraphics[width=0.85\linewidth]{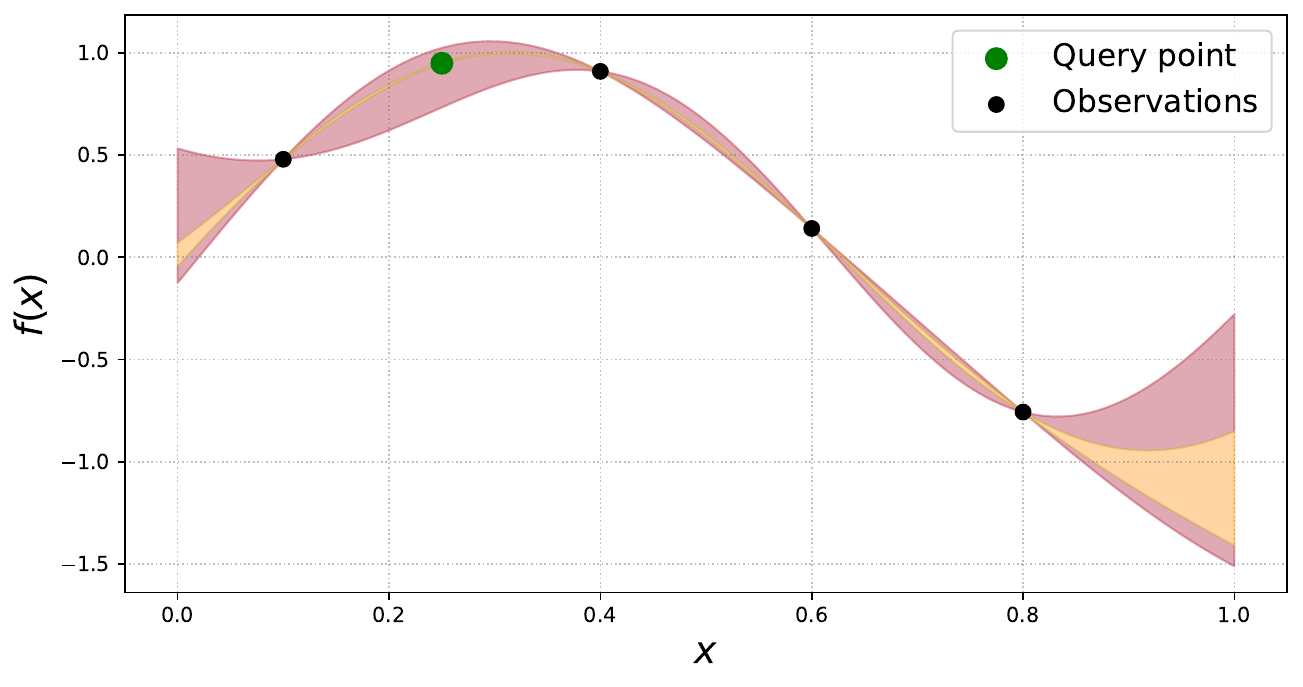}
\caption{The change in the global uncertainty when a candidate point is added to the observations. When the iteration $n=4$, we add a candidate point (green point) to the observations (black points) and measure the change of the global uncertainty (red region). The candidate point that maximizes the global uncertainty change will be selected as the next query point.}
\label{fig:GP}
\end{figure*}

Building on this, we propose a non-myopic strategy that considers the future impact of the uncertainty of all points from the entire input space $\mathcal{X}$, i.e., \textit{global information gain}. Specifically, given a candidate point, we evaluate the change of the \textit{global uncertainty} over the entire input space $\mathcal{X}$ when this point is added to the current observations. We aim to select the candidate point that maximizes this change (equivalently, minimizing the uncertainty the most) as the next query point. See Figure \ref{fig:GP}  for more illustration.

Given the candidate point $\tilde{\mathbf{x}}$ and current observations $\mathbf{x}_{1:n}$ at iteration $n$, we can formulate the global uncertainty over the input space $\mathcal{X}$ as 

\begin{equation}
\label{eq:global_uncer}
\tilde{s}(\mathbf{x})=\int_{\mathcal{X}} k(\mathbf{x},\mathbf{x})- \mathbf{k}_{n, \tilde{\mathbf{x}}}(\mathbf{x})^\top (\mathbf{K}_{n,\tilde{\mathbf{x}}}+\sigma^2_{\epsilon} \mathbf{I})^{-1}\mathbf{k}_{n,\tilde{\mathbf{x}}}(\mathbf{x})d\mathbf{x},
\end{equation}
where $(\mathbf{K}_{n,\tilde{\mathbf{x}}})_{ij}=k(\mathbf{x}_i,\mathbf{x}_{j})$ is the Gram matrix for all $n+1$ input data points, including the candidate point $\tilde{\mathbf{x}}$ and current observations $\mathbf{x}_{1:n}$; $\mathbf{k}_{n,\tilde{\mathbf{x}}}(\mathbf{x})=[k(\mathbf{x},\mathbf{x}_1),\dots,k(\mathbf{x},\mathbf{x}_{n}),k(\mathbf{x},\tilde{\mathbf{x}})]^\top$ calculates the kernel between the $\mathbf{x}$ and all $n+1$ input pairs.

\subsection{FigBO Algorithm}
Since maximizing the change in uncertainty is equivalent to minimizing the global uncertainty $\tilde{s}(\mathbf{x})$ and the first term in the integral of \eqref{eq:global_uncer} is a fixed constant for any candidate point, our objective can be defined as
\begin{equation}
\max_{\mathbf{x} \in \mathcal{X}} \bigg\{\Gamma(\mathbf{x})=\int_{\mathcal{X}} \mathbf{k}_{n,\tilde{\mathbf{x}}}(\mathbf{x})^\top (\mathbf{K}_{n,\tilde{\mathbf{x}}}+\sigma^2_{\epsilon} \mathbf{I})^{-1}\mathbf{k}_{n,\tilde{\mathbf{x}}}(\mathbf{x})d\mathbf{x} \bigg\},
\end{equation}
where the integral over the input space $\mathcal{X}$ is computationally intractable, we propose to approximate it through Monte Carlo (MC) method by uniformly sampling $L$ input points from $\mathcal{X}$. Therefore, the $\Gamma(\mathbf{x})$ can then be calculated as
\begin{equation}
\label{eq:Gamma}
\Gamma(\mathbf{x}) \approx \frac{1}{L} \sum_{l=1}^L \mathbf{k}_{n,l}(\mathbf{x})^\top (\mathbf{K}_{n,l}+\sigma^2_{\epsilon} \mathbf{I})^{-1}\mathbf{k}_{n,l}(\mathbf{x}).
\end{equation}
Since selecting each new candidate point as a query point requires fitting a GP model to obtain the results of \eqref{eq:Gamma}, this implies that we need to perform $L$ fittings to obtain $L$ GP models with different hyperparameter sets, resulting in extremely high computational complexity. Given the local smoothness of Gaussian process hyperparameters \cite{williams2006gaussian}, where small changes in the hyperparameter space have minimal impact on the model, we extend the hyperparameter set from the current observations $\mathbf{x}_{1:n}$ to include $n+1$ points by adding the candidate point $\tilde{\mathbf{x}}$. This allows us to achieve computational efficiency by using the Sherman-Morrison formula \cite{sherman1950adjustment} to iteratively update the inverse Gram matrix $(\mathbf{K}_{n,l}+\sigma^2_{\epsilon} \mathbf{I})^{-1}$ through rank-1 updates, reducing the per-sample complexity from $\mathcal{O}(n^3)$ for direct inversion to $\mathcal{O}(n^2)$.

To this end, we can combine $\Gamma(\mathbf{x})$ with any myopic acquisition function to equip it with look-ahead capability, thereby achieving a trade-off between future optimization benefits and simplicity of implementation. Formally, we choose the next query point $\mathbf{x}_{n+1}$ with the resulting strategy as
\begin{equation}
\label{eq:old_acq}
\mathbf{x}_{n+1} \in \mathrm{argmax}_{\mathbf{x}\in \mathcal{X}} \big( \alpha(\mathbf{x})+\Gamma(\mathbf{x}) \big),
\end{equation}
where $\alpha(\mathbf{x})$ denotes the myopic acquisition function such as EI, UCB, and PI. However, \eqref{eq:old_acq} indicates that the selection of the next query point throughout the optimization process is guided by global uncertainty, which causes the algorithm to overly focus on exploration while neglecting exploitation. Intuitively, the $\Gamma(\mathbf{x})$ helps the model quickly identify promising regions of the global optimum in the early stages. In the later stages, as the surrogate model has been sufficiently trained and exploitation becomes more important, we aim for the acquisition function to revert to its original form to balance exploration and exploitation. To achieve this, we introduce a coefficient $\lambda=\eta/n \in \mathbb{R}^+$ where $\eta$ serves as a hyperparameter that governs the decay rate of $\lambda$. This coefficient ensures that $\lambda$ diminishes to zero as the number of iterations $n$ grows, effectively reducing the impact of $\Gamma(\mathbf{x})$ over time. Thus, the next query point $\mathbf{x}_{n+1}$ can be chosen as
\begin{equation}
\label{eq:new_acq}
\mathbf{x}_{n+1} \in \mathrm{argmax}_{\mathbf{x}\in \mathcal{X}} \big( \alpha(\mathbf{x})+\lambda\Gamma(\mathbf{x}) \big).
\end{equation}

In the proposed FigBO, we achieve a balance between future optimization benefits and simplicity of implementation by combining acquisition strategy \eqref{eq:new_acq} with existing myopic acquisition functions. Additionally, we use coefficient $\lambda$ to control the exploration-exploitation trade-off throughout the optimization process, leading to more stable model performance. For clarity, we refer to the myopic acquisition function used in FigBO as the \textit{myopic base acquisition function}. The entire procedure is outlined in Algorithm \ref{alg:algorithm}.

\begin{algorithm}[tb]
    \caption{\texttt{FigBO}}
    \label{alg:algorithm}
    \textbf{Input}: Black-box function $f$, input space  $\mathcal{X}$, hyperparameter for controlling decay rate $\eta$, number of MC samples $L$, size of the initial design $M$,  max number of optimization iterations $N$\\
    % \textbf{Parameter}: Optional list of parameters\\
    \textbf{Output}: Optimized design $\mathbf{x}^*$
    \begin{algorithmic}[1] %[1] enables line numbers
        \STATE Initialize the data points: $\mathcal{D}_0 \leftarrow \{(\mathbf{x}_i,y_i)\}_{i=1}^M$
        \FOR{$\{n=1,2,\dots,N\}$}
        \STATE Fit a GP using the current observations $\mathcal{D}_n$
        \STATE Calculate the $\Gamma(\mathbf{x})$ by employing $n+1$ input points with $L$ MC samples as in \eqref{eq:Gamma}
        \STATE Obtain the next query point:  $\mathbf{x}_{\mathrm{new}} \leftarrow \arg \max_{\mathbf{x}\in \mathcal{X}} \big(\alpha(\mathbf{x})+\lambda \Gamma(\mathbf{x})\big)$, where $\lambda=\eta/n$
        \ENDFOR
        \RETURN $\mathbf{x}^* \leftarrow \mathrm{argmin}_{(\mathbf{x}_i,y_i)\in \mathcal{D}_N}y_i$
        % \WHILE{condition}
        % \STATE Do some action.
        % \IF {conditional}
        % \STATE Perform task A.
        % \ELSE
        % \STATE Perform task B.
        % \ENDIF
        % \ENDWHILE
        % \STATE \textbf{return} solution
    \end{algorithmic}
\end{algorithm}

\section{Theoretical Analysis}
In this section, we present the convergence rate for FigBO. To concretize the research problem, we choose EI as our myopic base acquisition function, as it is currently the most widely used myopic acquisition function. Our work follows the same assumptions as those introduced by \cite{bull2011convergence} and \cite{hvarfner2022pi}. These assumptions hold for widely used kernels, such as the Matérn kernel and the Gaussian kernel. The latter can be viewed as a special case of the Matérn kernel when the smoothness parameter $\nu \rightarrow \infty$, where $\nu$ controls the smoothness of functions in the GP prior. In this convergence analysis, the theoretical results hold regardless of whether the kernel hyperparameters, including the length scales $\ell$ and the signal variance $\sigma^2$, are fixed or learned using Maximum Likelihood Estimation (MLE). Given the function $f$ with a symmetric positive-definite kernel $K_{\ell}$, we define the loss associate with the ball $B_R$ in the reproducing kernel Hilbert space (RKHS) $\mathcal{H}_{\ell}(\mathcal{X})$ as
\begin{equation}
\label{eq:loss}
\mathcal{L}_n(u, \mathcal{H}_{\ell}(\mathcal{X}),R)\overset{\Delta}{=} \underset{\|f\|_{\mathcal{H}_{\ell}(\mathcal{X})} \leq R}{\mathrm{sup}} \mathbb{E}_{f}^{u} [f(\mathbf{x}^*_n)-\min f],
\end{equation}
where $u$ is the strategy for choosing $\mathbf{x}_n$ and converges to the optimum at rate $r_n$ if $\mathcal{L}_n(u, \mathcal{H}_{\ell}(\mathcal{X}),R)=O(r_n)$ for all $R>0$. It should be emphasized that $u$ cannot vary with $R$; the strategy must achieve this rate independently of any prior information about $\|f\|_{\mathcal{H}_{\ell}(\mathcal{X})}$ \cite{bull2011convergence}. 

Based on \eqref{eq:loss}, we will next present the expected regret bound and convergence rate of the proposed optimization strategy $EI_{\Gamma,n}$, as well as their relationship to those of the vanilla optimization strategy $EI_n$. The full proofs are provided in the Appendix.

\begin{theorem}
\label{theorem1}
For the $\mathcal{D}_n$, $\mathcal{H}_{\ell}(\mathcal{X})$, $R$ defined above, the loss $\mathcal{L}_n$ of $EI_{\Gamma,n}$ with a compact set $\mathcal{X}$ at iteration $n$ can satisfy
\begin{equation}
\begin{aligned}
\mathcal{L}_n(\mathrm{EI}_{\Gamma,n},\mathcal{D}_n,\mathcal{H}_{\ell}(\mathcal{X}),R) &\leq \mathcal{L}_n(\mathrm{EI}_{n},\mathcal{D}_n,\mathcal{H}_{\ell}(\mathcal{X}),R)+C'\\
& \leq \frac{\tau(R/\sigma)}{\tau(-R/\sigma)} \big(2Rm^{-1}+(R+\sigma)Cm^{-(\nu \wedge 1)/d}(\log m)^\beta\big)+C',
\end{aligned}
\end{equation}
where $C'=\frac{\tau(R/\sigma)}{\tau(-R/\sigma)} \underset{x\in \mathcal{X}}{\max} \lambda \Gamma(\mathbf{x})$.
\end{theorem}
\begin{remark}
Theorem \ref{theorem1} demonstrates that the difference in expected regret bounds between $EI_{\Gamma,n}$ and the vanilla $EI_n$ is solely determined by the term $C'$, which is closely related to the global information gain. 
\end{remark}

\begin{corollary}
\label{corollary1}
If $n \rightarrow \infty$, the loss of $EI_{\Gamma,n}$ becomes asymptotically equivalent to the loss of the vanilla $EI_n$:
\begin{equation}
\mathcal{L}_n(\mathrm{EI}_{\Gamma,n},\mathcal{D}_n,\mathcal{H}_{\ell}(\mathcal{X}),R) \sim \mathcal{L}_n(\mathrm{EI}_{n},\mathcal{D}_n,\mathcal{H}_{\ell}(\mathcal{X}),R),
\end{equation}
we thus obtain the convergence rate $\mathcal{L}_n(\mathrm{EI}_{\Gamma,n},\mathcal{D}_n,\mathcal{H}_{\ell}(\mathcal{X}),R)=\mathcal{O}(n^{-(\nu \wedge 1)/d}(\log n)^\beta)$, which is consistent with the convergence rate of the vanilla $\mathcal{L}_n(\mathrm{EI}_{n},\mathcal{D}_n,\mathcal{H}_{\ell}(\mathcal{X}),R)$ presented in \cite{bull2011convergence}.
\end{corollary}
\begin{remark}
The result of Corollary \ref{corollary1} shows that, as the number of iterations $n$ tends to infinity, the convergence performance of $EI_{\Gamma,n}$ is close to that of the vanilla $EI_n$, indicating that the introduction of the $\Gamma(\mathbf{x})$ does not harm the convergence rate of the vanilla $EI_n$. This conclusion is reasonable from the perspective of the entire optimization process. In the early stages of optimization, $EI_{\Gamma,n}$ places more emphasis on the variation of global information gain to identify promising regions, accelerating the optimization process and improving model performance. However, in the later stages of optimization (when the number of iterations $n$ becomes very large), the influence of global information gain on guiding the model gradually weakens, and the model progressively reverts to the vanilla $EI_n$, ensuring asymptotic convergence performance.
\end{remark}

\section{Experiment}
In this section, we examine the performance of FigBO on a suite of diverse tasks, including GP prior samples, synthetic test functions, and multilayer perceptron (MLP) for classification tasks. Unless specifically stated otherwise, we use EI as our default myopic base acquisition function to implement FigBO. We compare FigBO against other popular myopic acquisition functions, including EI, UCB, and PI, as well as state-of-the-art non-myopic acquisition functions such as PES, MES, and JES. The hyperparameter of UCB is chose based on Theorem 2 in \cite{srinivas2012information}, while the hyperparameter of PI is set equal to the observation noise $\sigma_{\epsilon}$. All the synthetic experiments are executed over $200$ iterations, whereas the MLP classification tasks are evaluated over $100$ iterations. The number of repetitions for the synthetic experiments and MLP tasks is set to 20 and 50, respectively. To mitigate the impact of varying the total number of iterations $N$ on model performance across different experiments, we set $\eta$ to $N/10$ in all experiments. For a fair comparison, we fix the number of MC samples to $100$ for FigBO, PES, MES, and JES. More details about experimental design can be found in the Appendix.

\subsection{GP Prior Samples}
\begin{figure}[t]
\centering
\includegraphics[width = 0.48\textwidth]{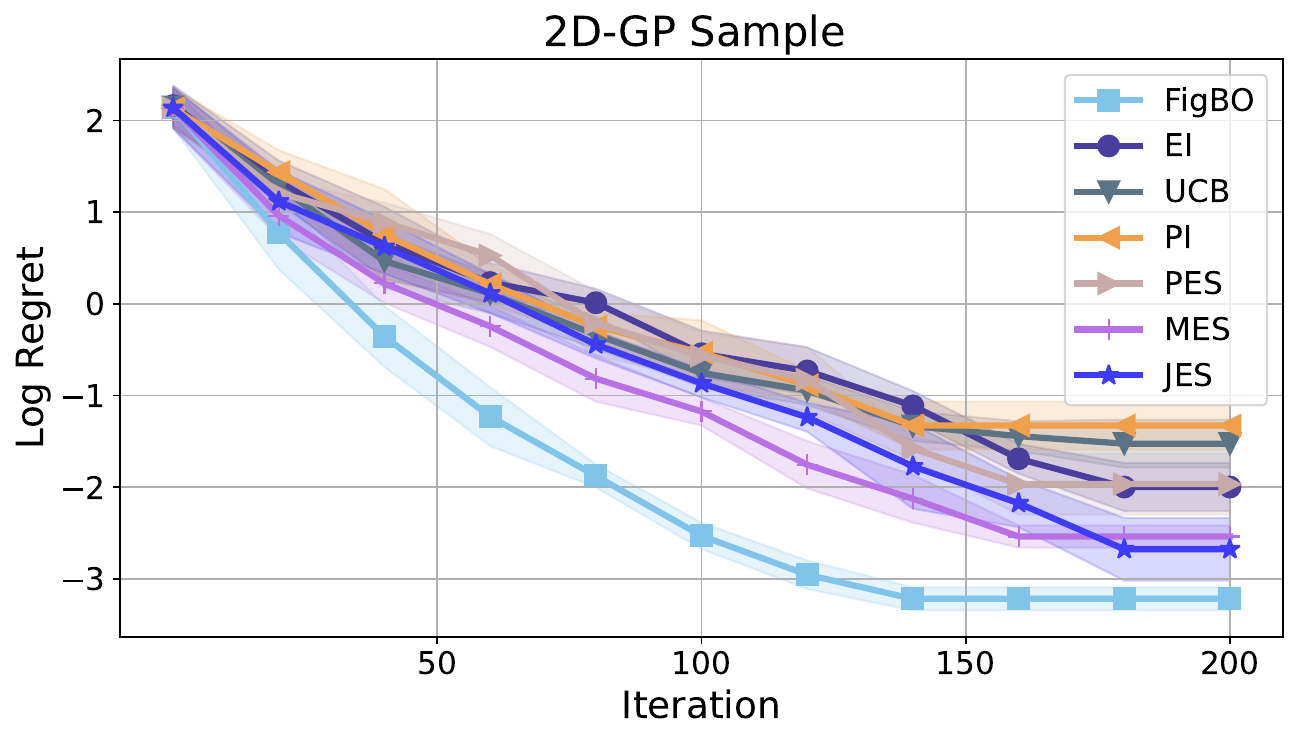}
\includegraphics[width =0.48\textwidth]{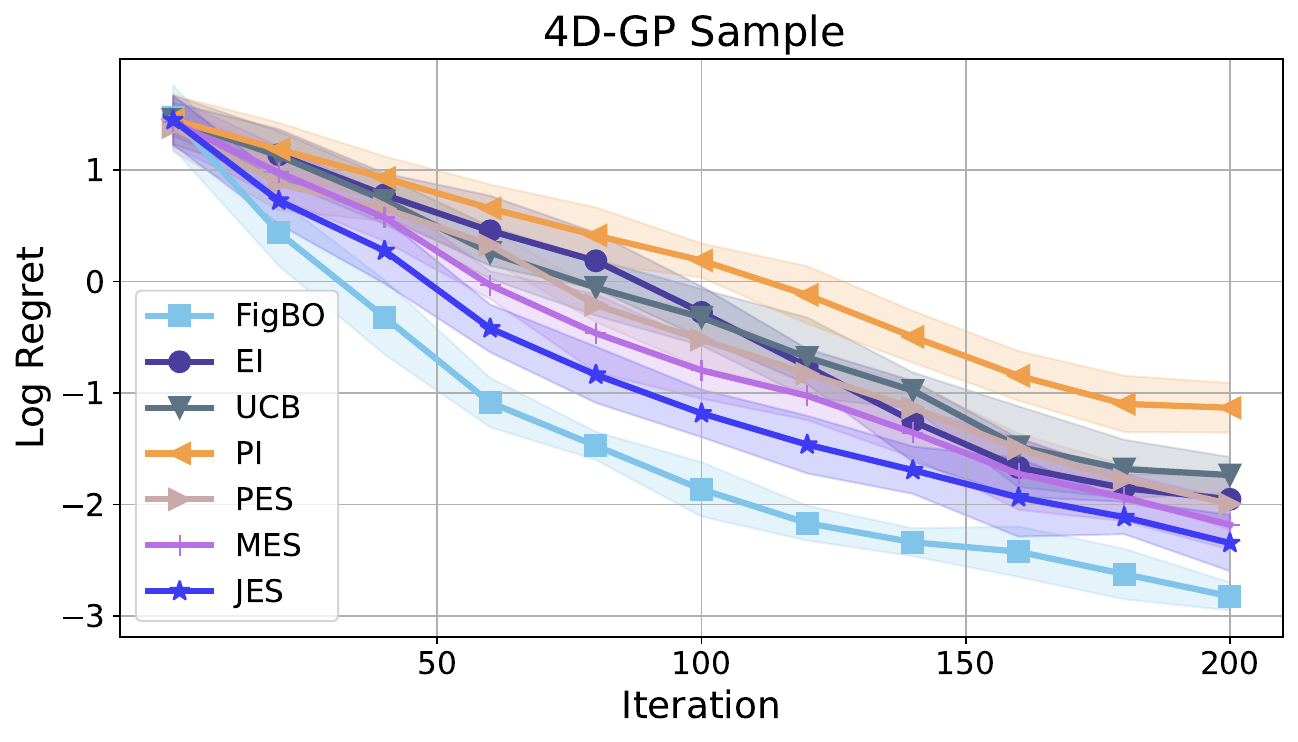}
\includegraphics[width =0.48\textwidth]{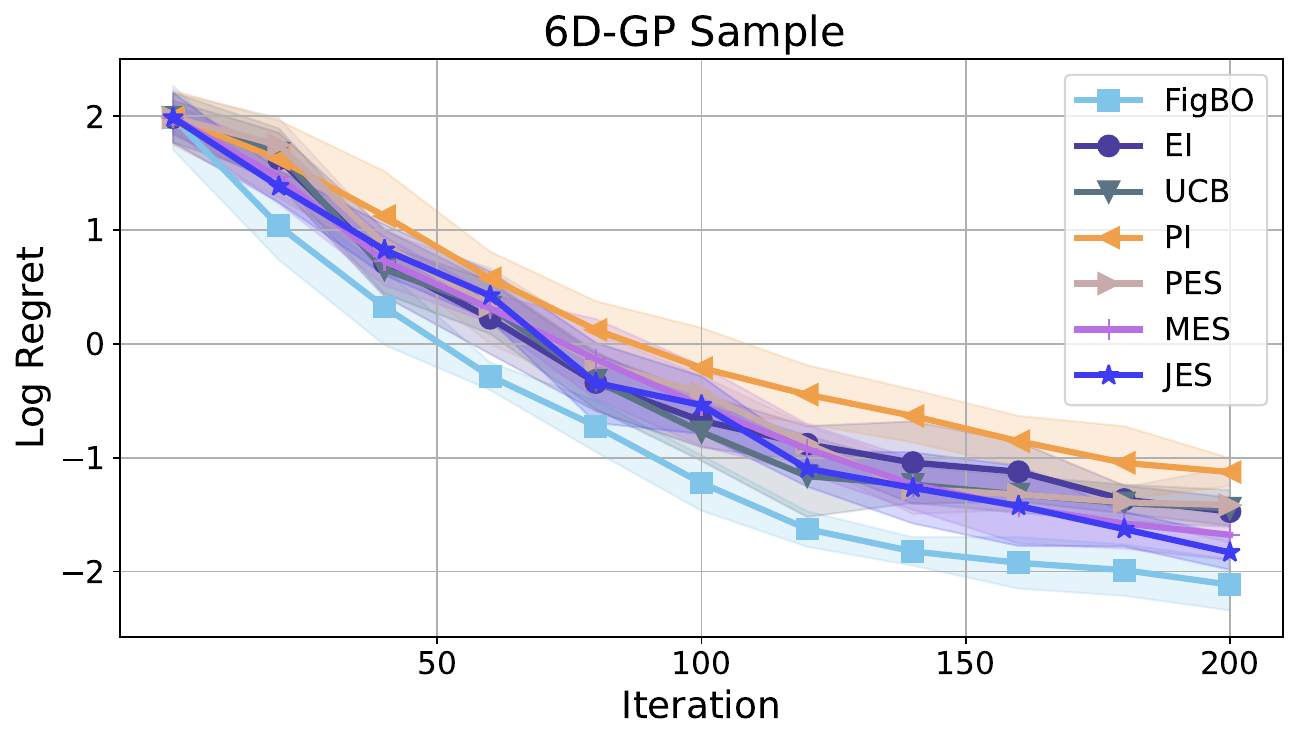}
\includegraphics[width =0.48\textwidth]{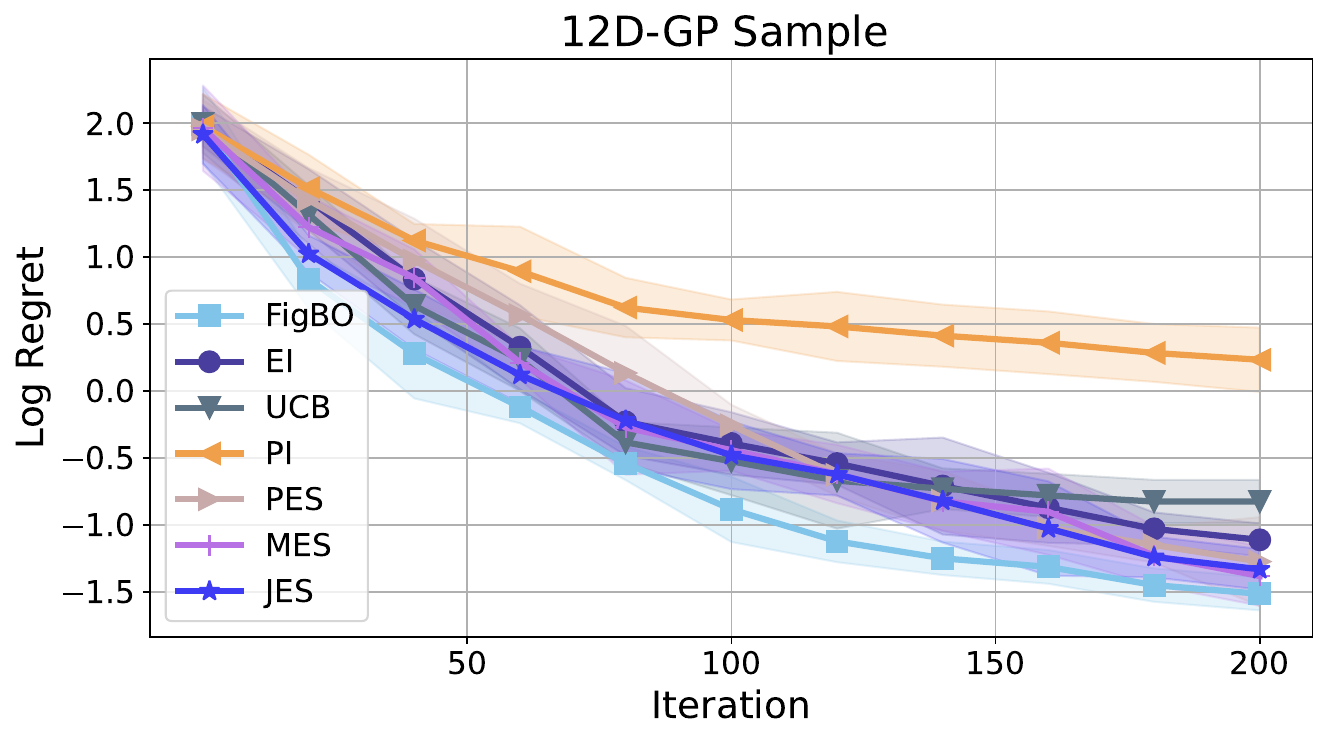}
\caption{Average log regret of all baselines on GP prior sample tasks across different dimensions (2D, 4D, 6D, 12D) over 20 repetitions. The average log regret and its standard errors are displayed for each acquisition function.}
\label{fig:prior}
\end{figure}

To evaluate the performance of each algorithm in an ideal setting, we generate synthetic data by sampling from a GP prior across four different input dimensions: 2D, 4D, 6D, and 12D \cite{hvarfner2022joint}. We introduce Gaussian noise with a standard deviation of $0.1$, and the function outputs are scaled to approximately range within $[-10,10]$. Since the surrogate model is also a GP, it perfectly matches the data-generating process, ensuring a fair comparison across all methods. The log regret results are recorded in Fig. \ref{fig:prior}, from which we can see that FigBO consistently outperforms other baselines in GP prior tasks across different dimensions (2D, 4D, 6D, and 12D), demonstrating its strong adaptability and robustness. Notably, FigBO exhibits significantly faster convergence in 2D, 4D, and 6D tasks compared to other methods, which can be attributed to its ability to leverage global information gain to identify promising search regions more efficiently. In the 12D task, however, this advantage weakens due to the increased difficulty of efficiently identifying promising regions in high-dimensional spaces.

We also compare the runtime for all acquisition functions on the GP prior sample tasks.  The runtime for each acquisition function is measured as the time elapsed after the GP hyperparameters are sampled and the GP covariance matrix is inverted, up to the point where the next query is selected. This includes the time spent on setting up the acquisition function and optimizing it, but excludes the time required to evaluate the black-box function itself. The results in Table \ref{tab:runtime} show that FigBO is faster than non-myopic acquisition functions, particularly PES. Compared to myopic acquisition functions, FigBO's additional term $\Gamma(\mathbf{x})$ introduces a slight computational overhead, but this extra time is minimal and is outweighed by its performance benefits.

\begin{table}[t]
\renewcommand{\arraystretch}{1.0}
\caption{Runtime comparison for GP prior sample tasks with different dimensions.}
\label{tab:runtime}
\centering
% \resizebox{\linewidth}{!}{
\begin{tabular}{cccccccc}
\toprule
 Task & FigBO & EI & UCB & PI & PES & MES & JES\\
 \midrule
 2D & 0.83±0.09 & 0.34±0.05 & 0.38±0.08 & 0.43±0.05& 15.77±3.22 & 1.13±0.23 & 1.43±0.22\\ 
 4D & 1.05±0.11 & 0.42±0.08 & 0.46±0.12 & 0.51±0.09 & 31.73±7.61 & 1.26±0.28 & 1.55±0.26\\
 6D & 1.13±0.15 & 0.48±0.12 & 0.51±0.16 & 0.54±0.13 & 57.88±8.36 & 1.38±0.31 & 1.62±0.31\\
 12D & 1.36±0.21 & 0.69±0.23 & 0.73±0.22 & 0.76±0.25 & 151.58±21.43 & 1.61±0.34 & 1.95±0.35\\
\bottomrule
\end{tabular}
\end{table}

\subsection{Synthetic Test Functions}
\begin{figure}[t]
\centering
\includegraphics[width = 0.32\textwidth]{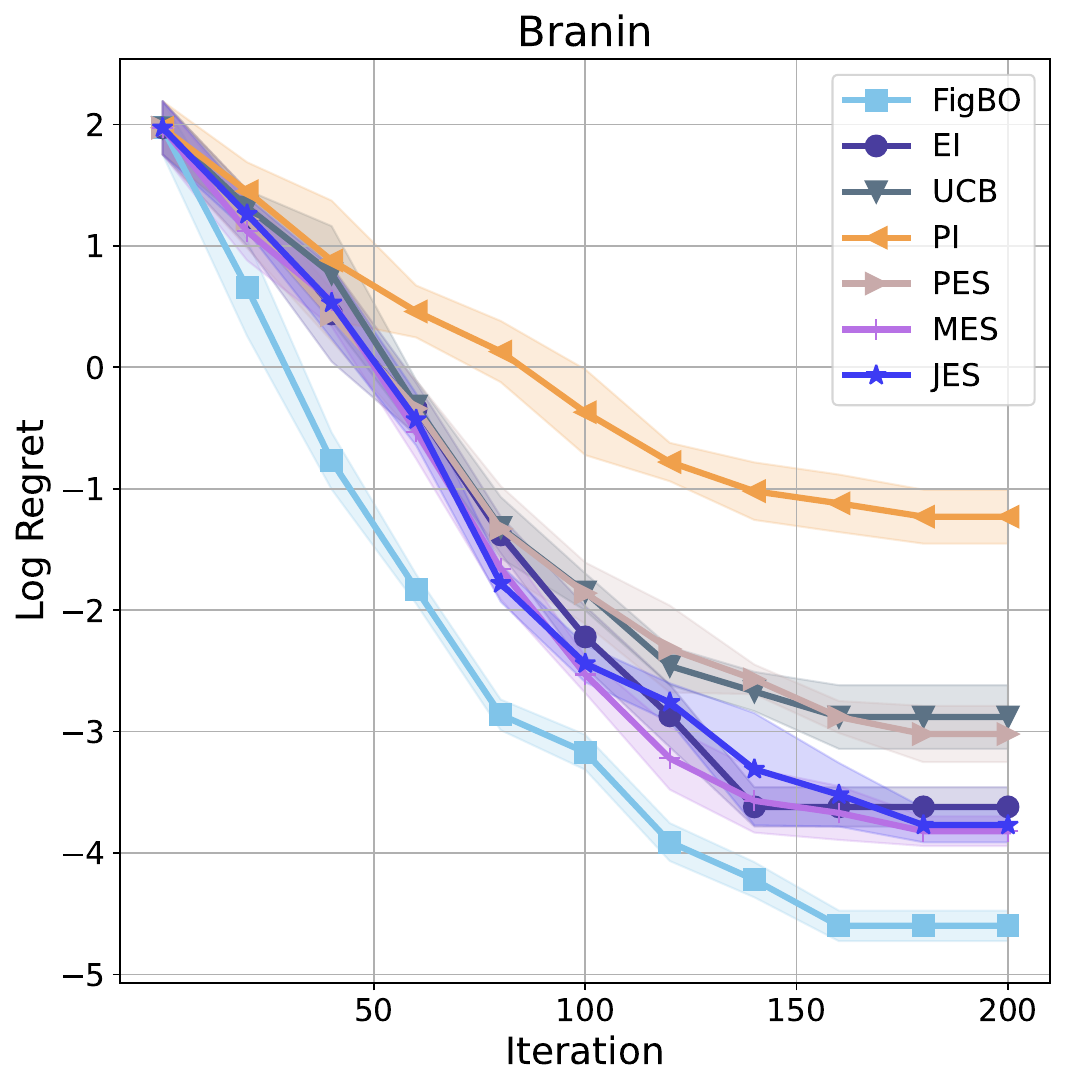}
\includegraphics[width =0.32\textwidth]{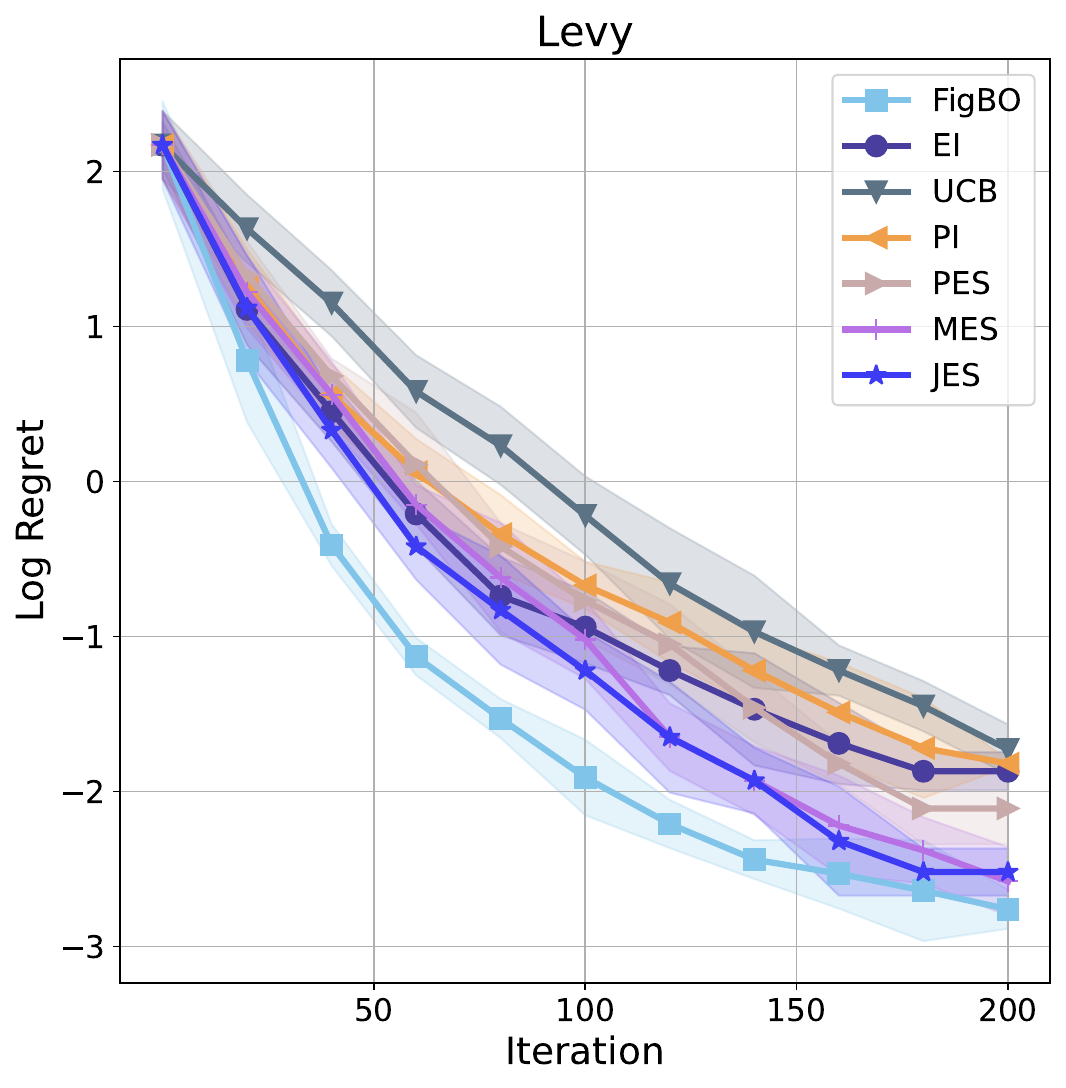}
\includegraphics[width =0.32\textwidth]{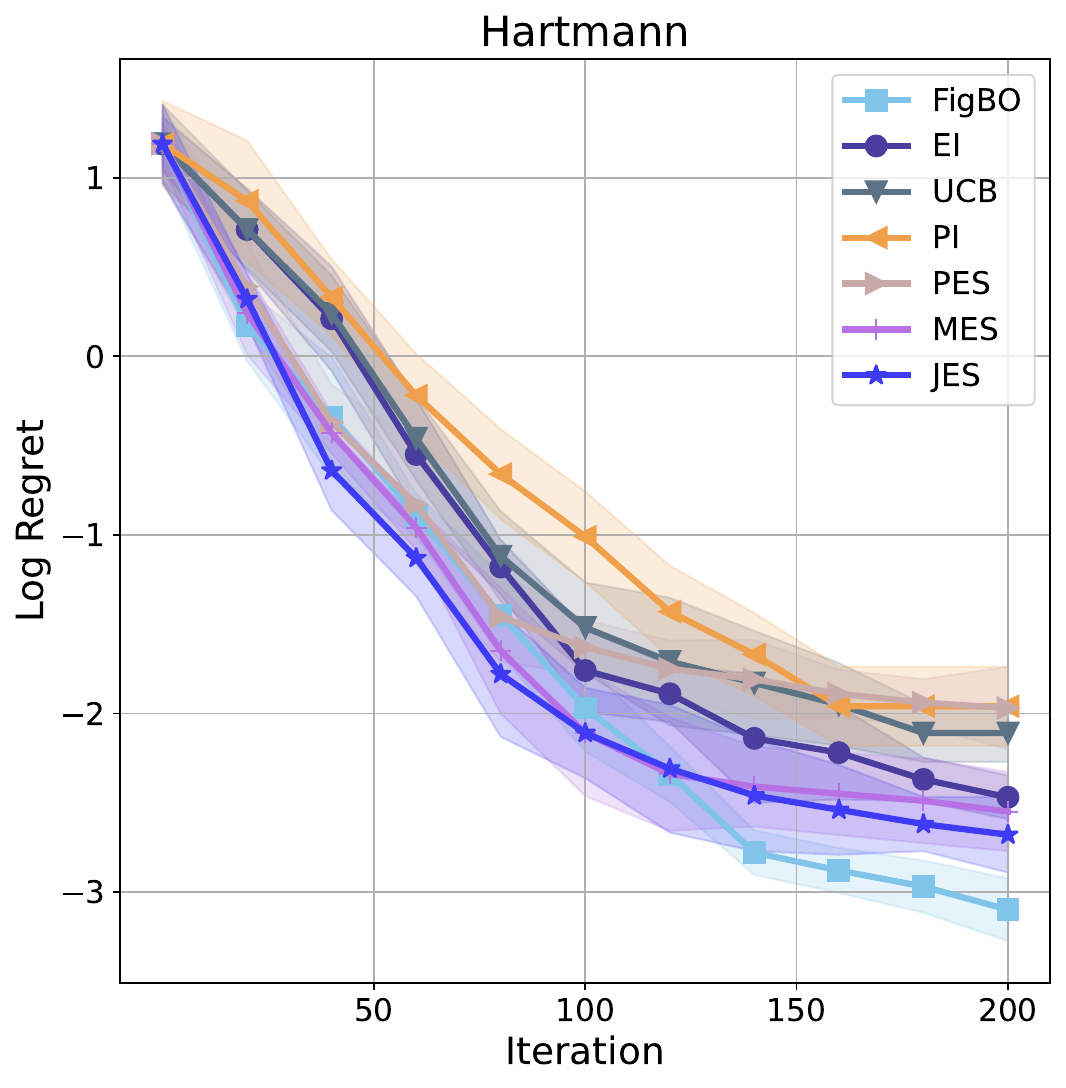}
\caption{Average log regret of all baselines on three different synthetic functions over 20 repetitions. The average log regret and its standard errors are displayed for each acquisition function.}
\label{fig:synthetic}
\end{figure}

Next, we test the performance of FigBO on three popular standard synthetic functions: the 2-dimensional Branin function, the 4-dimensional Levy function, and the 6-dimensional Hartmann function. The results for log regret are shown in Fig. \ref{fig:synthetic}. We can observe that FigBO dominates the other methods in terms of average log regret on all synthetic functions. Additionally, FigBO demonstrates significantly faster convergence on the 2-dimensional Branin and 4-dimensional Levy functions compared to other baselines. On the 6-dimensional Hartmann function, FigBO's convergence speed is only slightly slower than that of MES and JES, while still maintaining competitive performance.

\subsection{MLP Classification Tasks}
\begin{figure}[t]
\centering
\includegraphics[width = 0.48\textwidth]{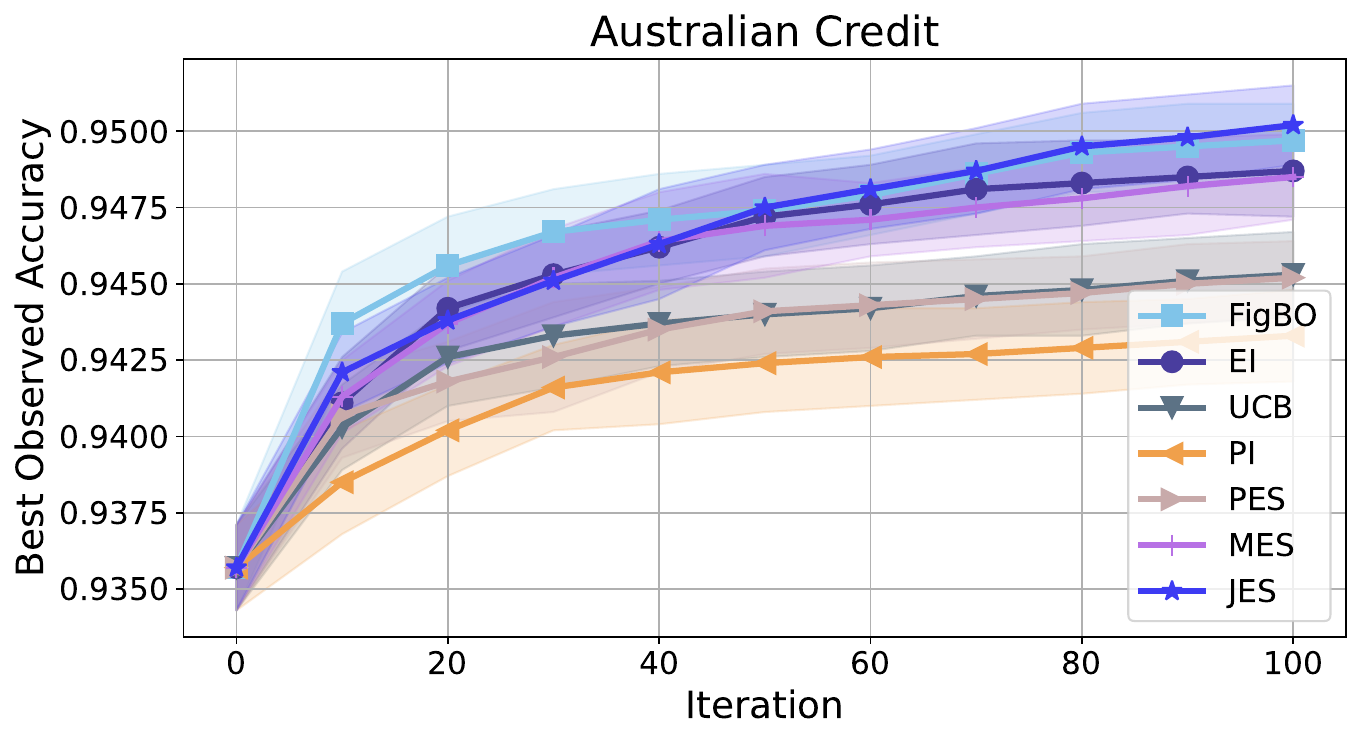}
\includegraphics[width =0.48\textwidth]{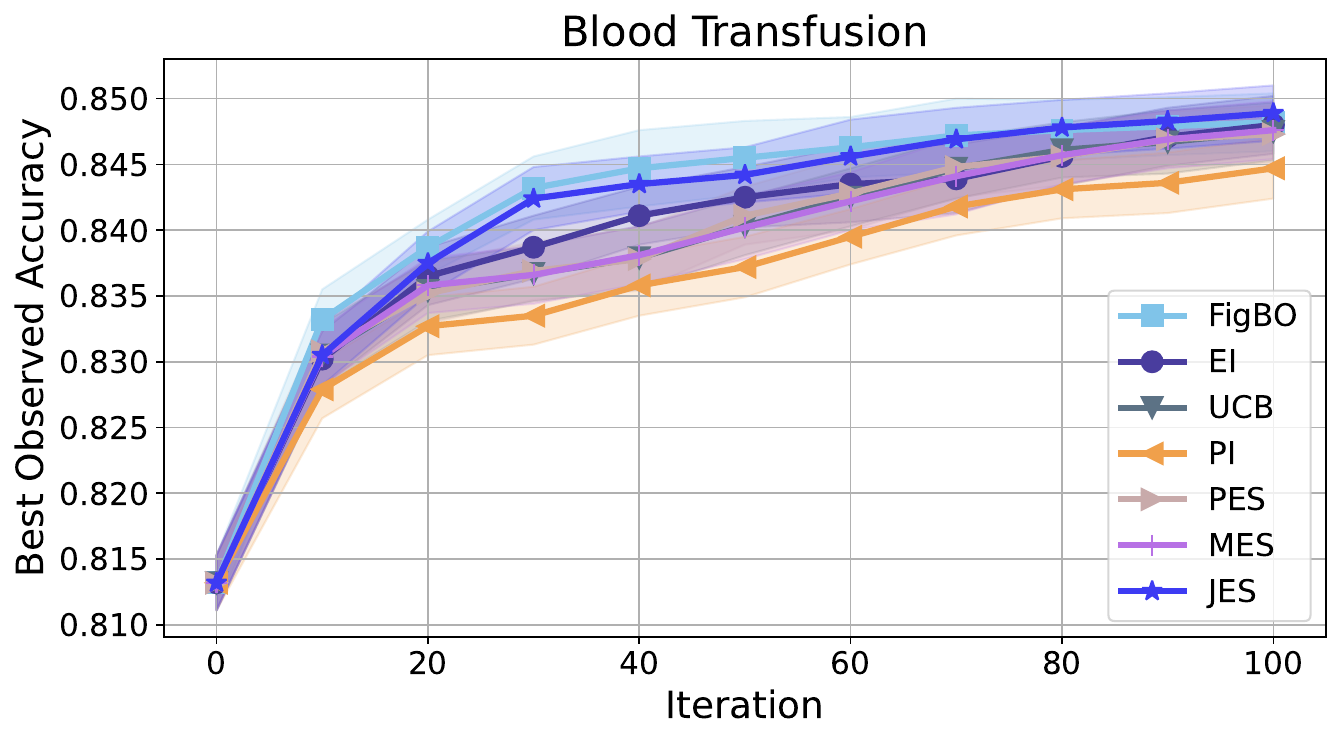}
\includegraphics[width =0.48\textwidth]{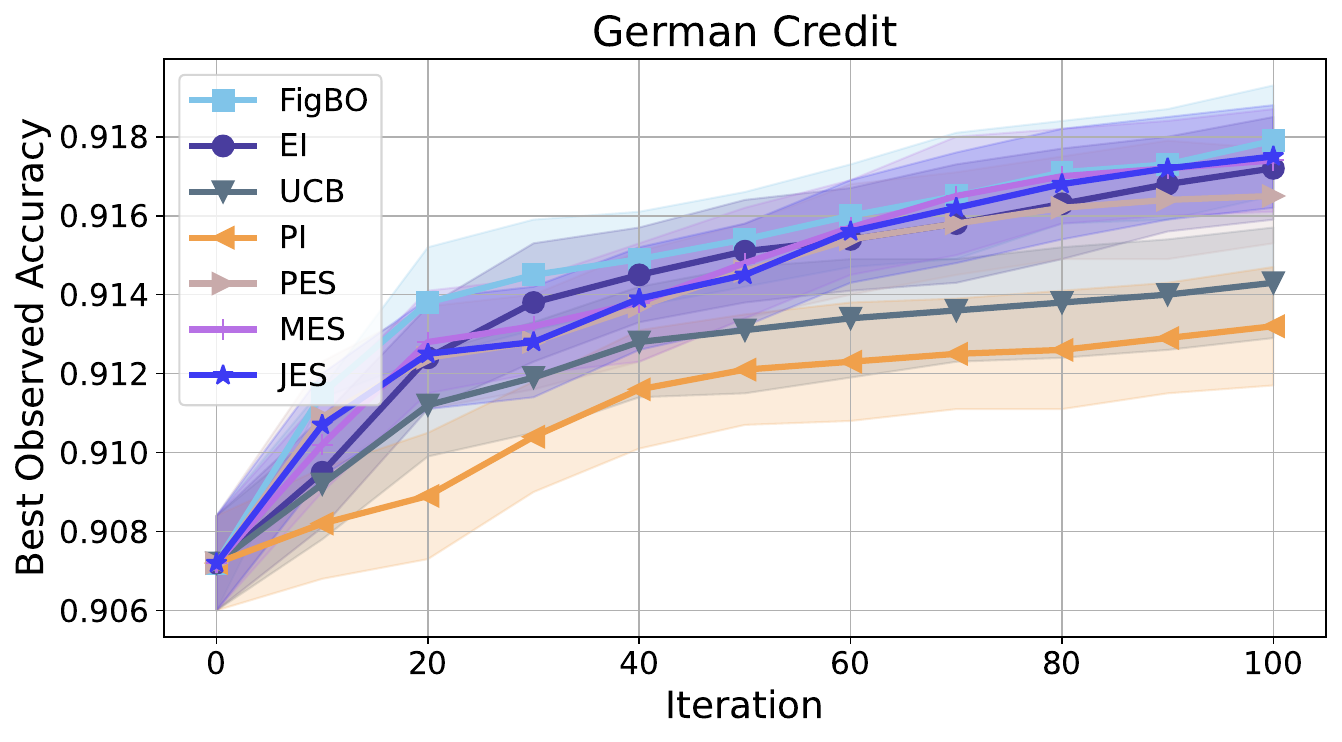}
\includegraphics[width =0.48\textwidth]{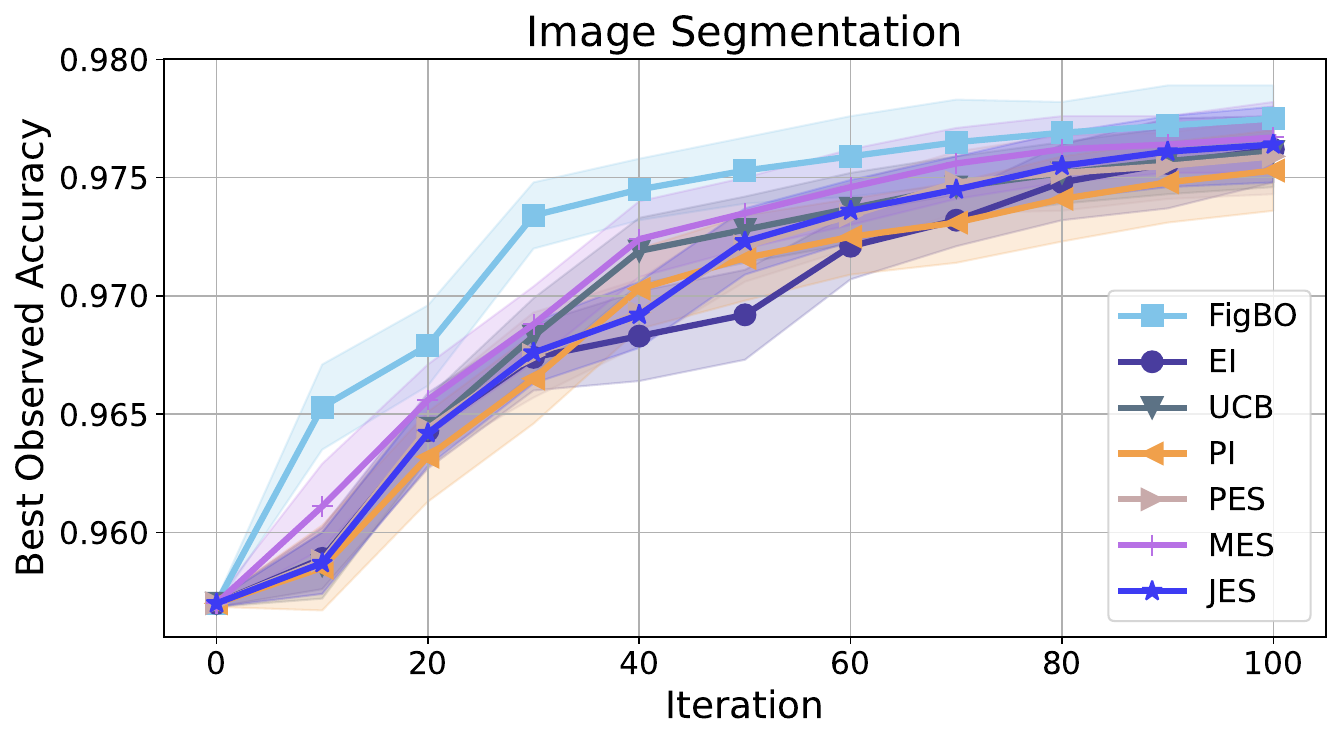}
\caption{Best observed accuracy of all baselines on four different MLP classification tasks over 50 repetitions. The average accuracy and its standard errors are displayed for each acquisition function.}
\label{fig:MLP}
\end{figure}

In this section, we test our FigBO algorithm on four different MLP classification tasks, which are included in the OpenML task repository, and the hyperparameter optimization benchmark is made available via the HPOBench suite \cite{eggensperger2021hpobench}. There are 4 hyperparameters in these tasks that need to be optimized. The main results are presented in Fig. \ref{fig:MLP}, we can observe that both FigBO and JES consistently outperform other methods across all four tasks. Moreover, FigBO demonstrates significantly faster convergence in all tasks, particularly in the image segmentation task. We also notice that PI and UCB exhibit unsatisfactory performance in the Australian credit and the German credit tasks, achieving significantly lower accuracy compared to other methods.

\subsection{Universality of FigBO}
\begin{figure}[t]
\centering
\includegraphics[width = 0.32\textwidth]{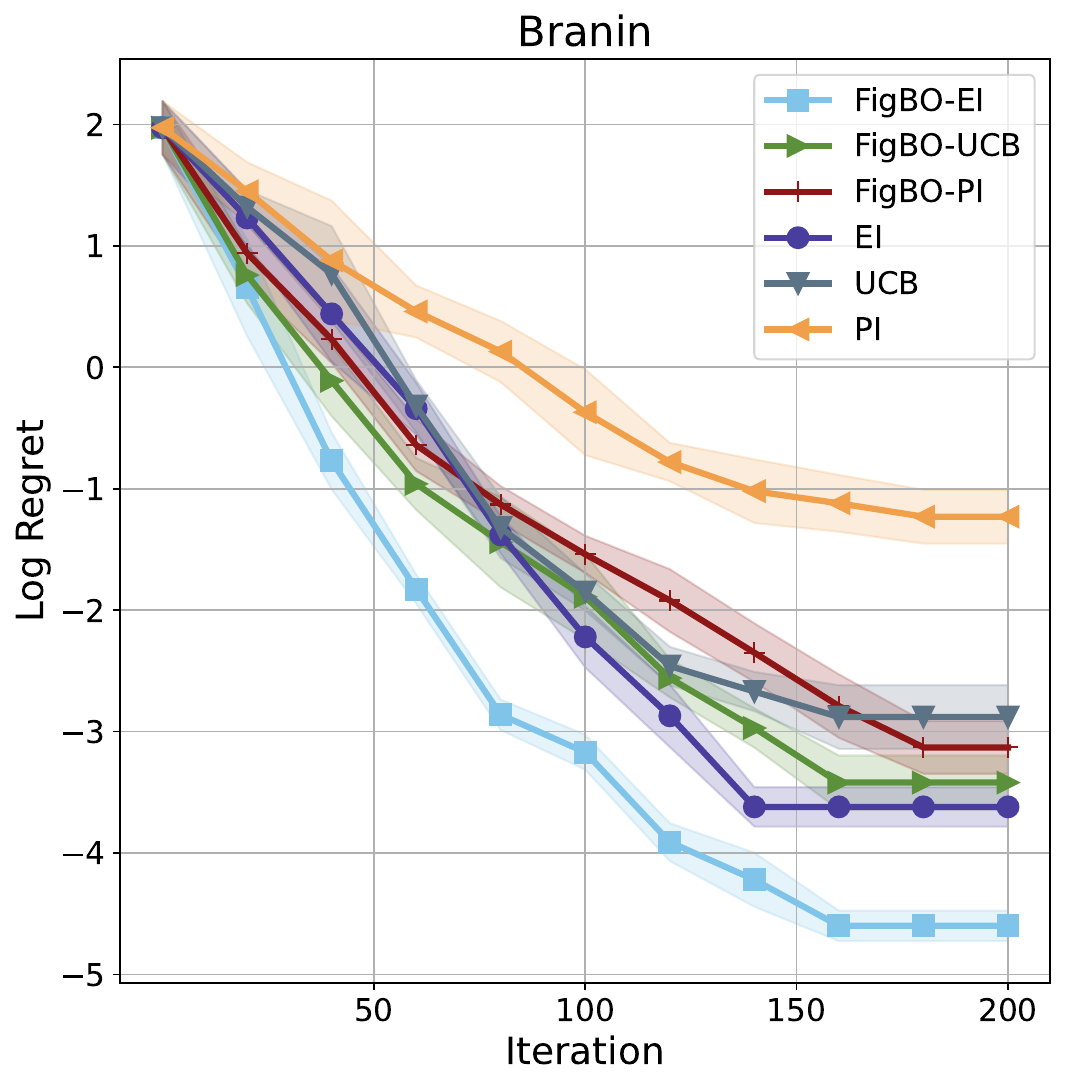}
\includegraphics[width =0.32\textwidth]{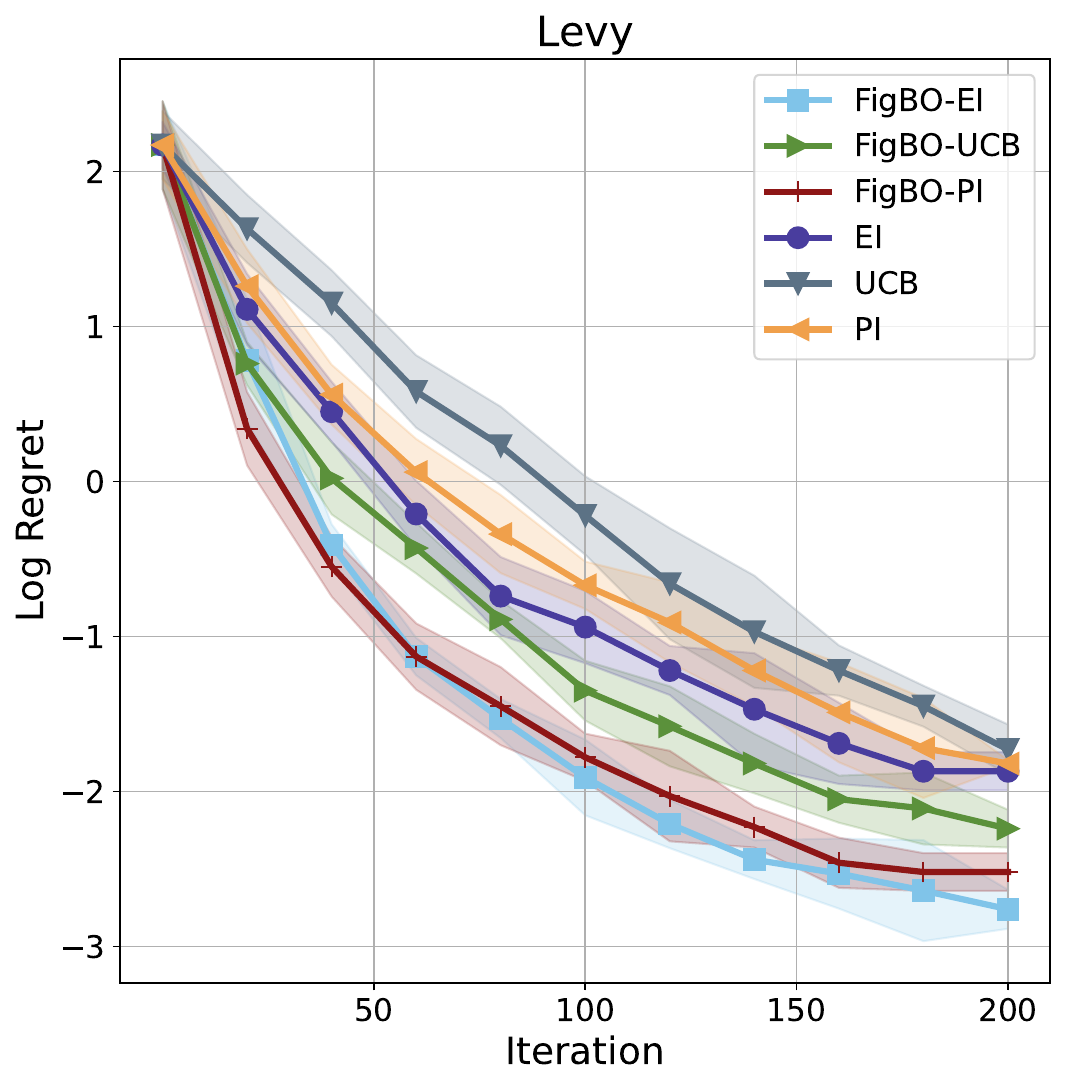}
\includegraphics[width =0.32\textwidth]{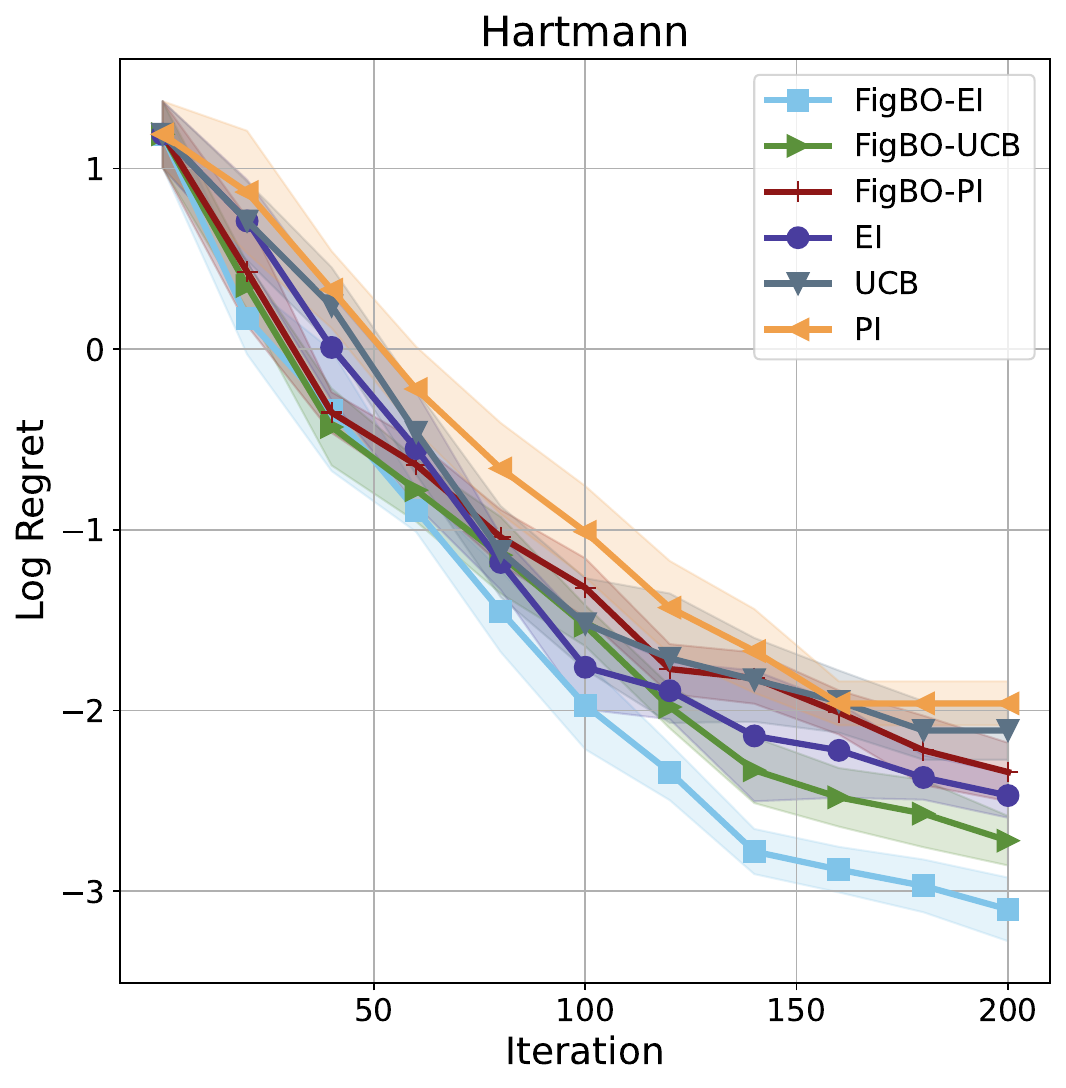}
\caption{Average log regret of diverse myopic acquisition functions and their FigBO variants on three different synthetic functions over 20 repetitions. The average log regret and its standard errors are displayed for each acquisition function.}
\label{fig:universality}
\end{figure}

To verify the universality of FigBO when combined with different myopic base acquisition functions, we extend our evaluation beyond EI to include UCB and PI as myopic base acquisition functions, referring to these variants as FigBO-EI, FigBO-UCB, and FigBO-PI. Since the output values of UCB are typically on the same scale as the objective function, we apply an affine transformation to the observations $\{y_i\}_i^n$ by substracting the current best value $y_n^*$ to ensure scale invariance \cite{hvarfner2022pi}. The results in Fig. \ref{fig:universality} showed that the FigBO variants of the three myopic acquisition functions achieve better log regret and faster convergence across all datasets compared to their original counterparts. Especially for the 2-dimensional Branin function, the improvement is particularly significant for PI and EI. These results highlight that incorporating global information gain to equip myopic acquisition functions with look-ahead capability plays a positive role in enhancing their performance.

\subsection{Effect of MC Samples}
\begin{figure}[t]
\centering
\includegraphics[width = 0.32\textwidth]{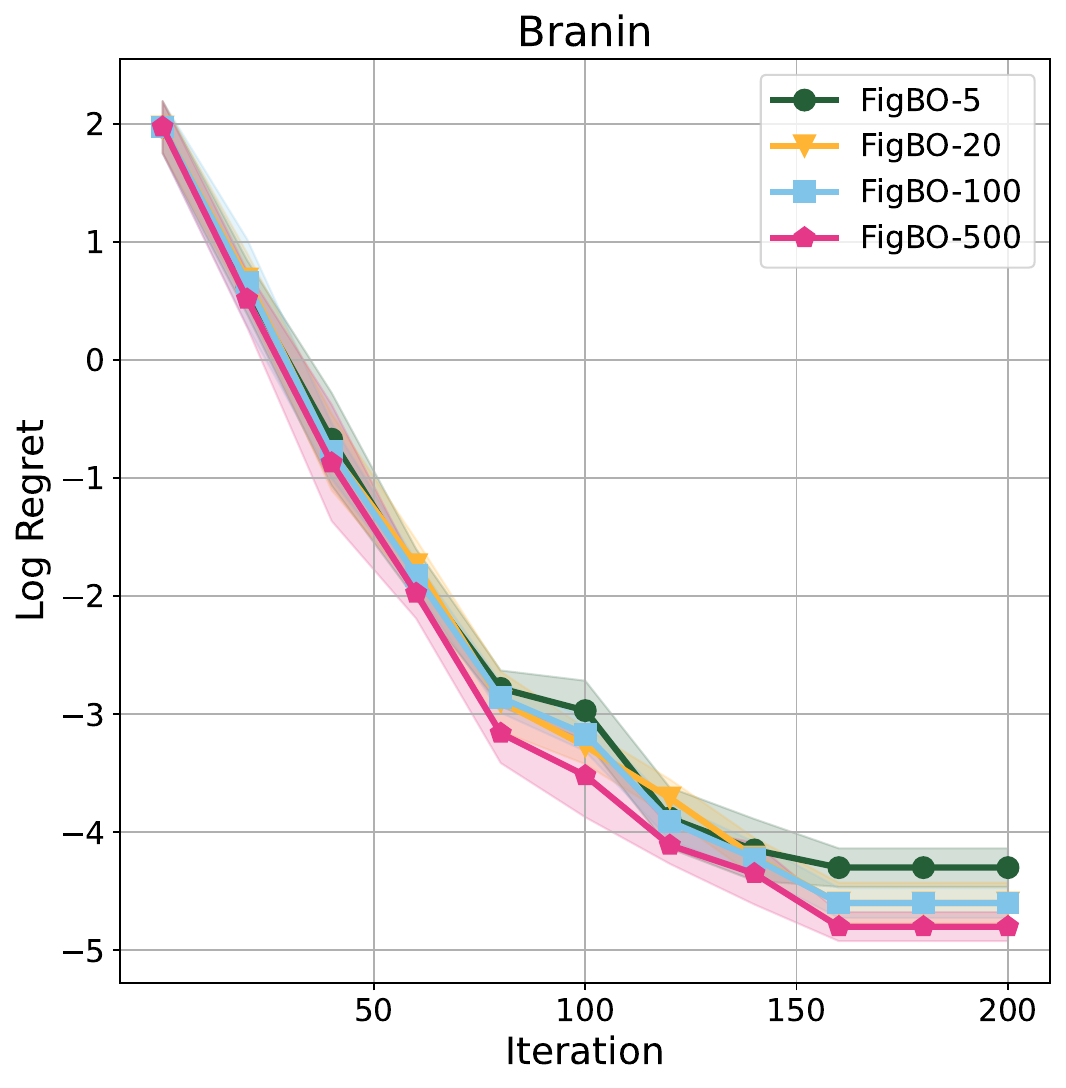}
\includegraphics[width =0.32\textwidth]{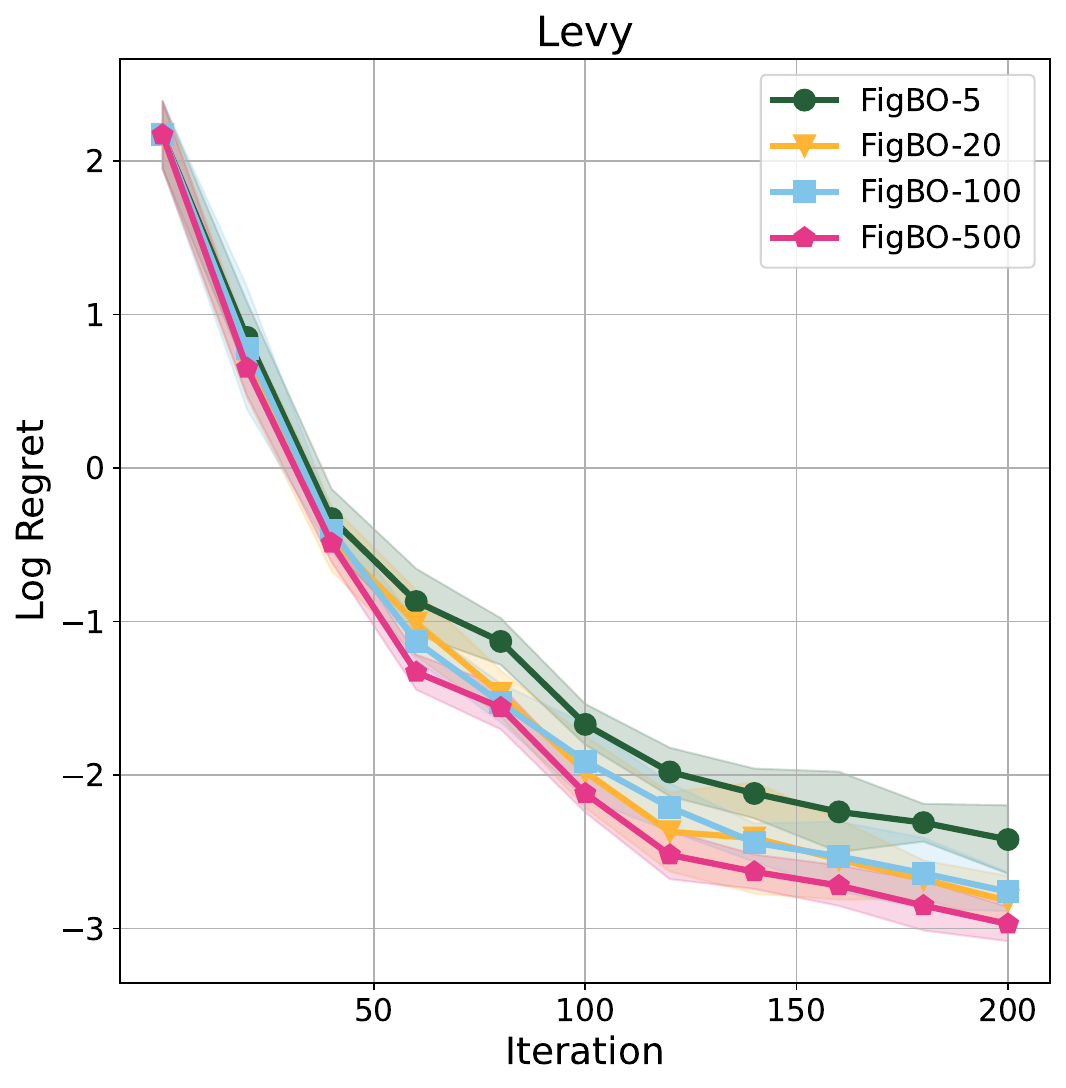}
\includegraphics[width =0.32\textwidth]{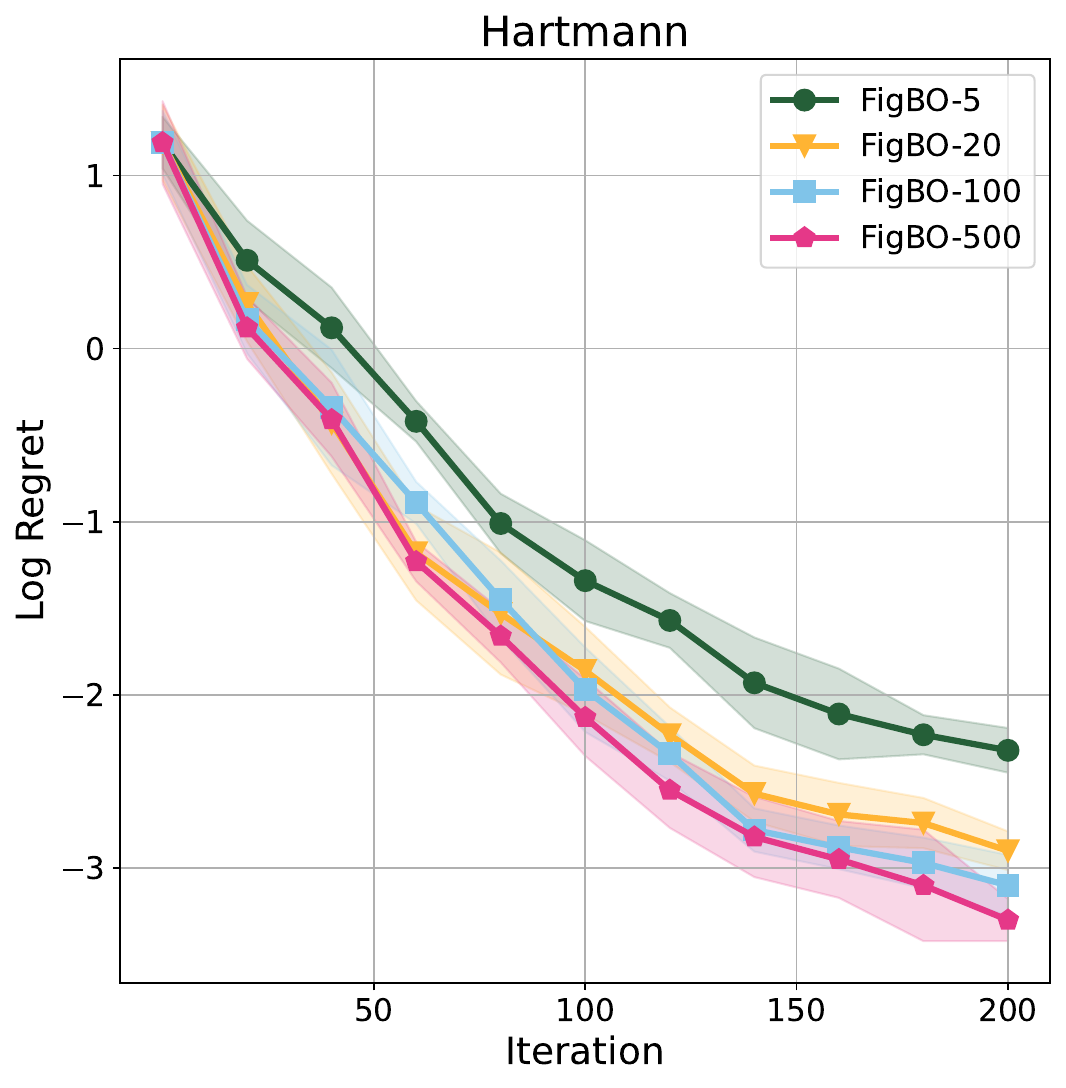}
\caption{Average log regret of FigBO with varying MC samples on three different synthetic functions over 20 repetitions. The average log regret and its standard errors are displayed for all tasks.}
\label{fig:MC}
\end{figure}

The number of MC samples significantly impacts the performance of FigBO, particularly in high-dimensional tasks. While a larger number of MC samples improves the accuracy of the integral approximation, it also increases computational complexity, making it crucial to strike a balance between precision and efficiency. To investigate this trade-off, we conduct experiments on three synthetic test functions by varying the number of MC samples over $\{5, 20, 100, 500\}$. The results in Fig. \ref{fig:MC} showed that n low-dimensional tasks, such as the 2-dimensional Branin function, the number of samples has a marginal impact on model performance. However, as the dimensionality increases, a larger number of samples leads to significantly better performance, as evidenced by the notable difference between using 5 samples and 500 samples in the 6-dimensional Hartmann function. The main reason is that high-dimensional spaces require more samples to accurately approximate $\Gamma(\mathbf{x})$ compared to low-dimensional spaces.

\subsection{Effect of Hyperparameter $\eta$}
\begin{figure}[t]
\centering
\includegraphics[width = 0.32\textwidth]{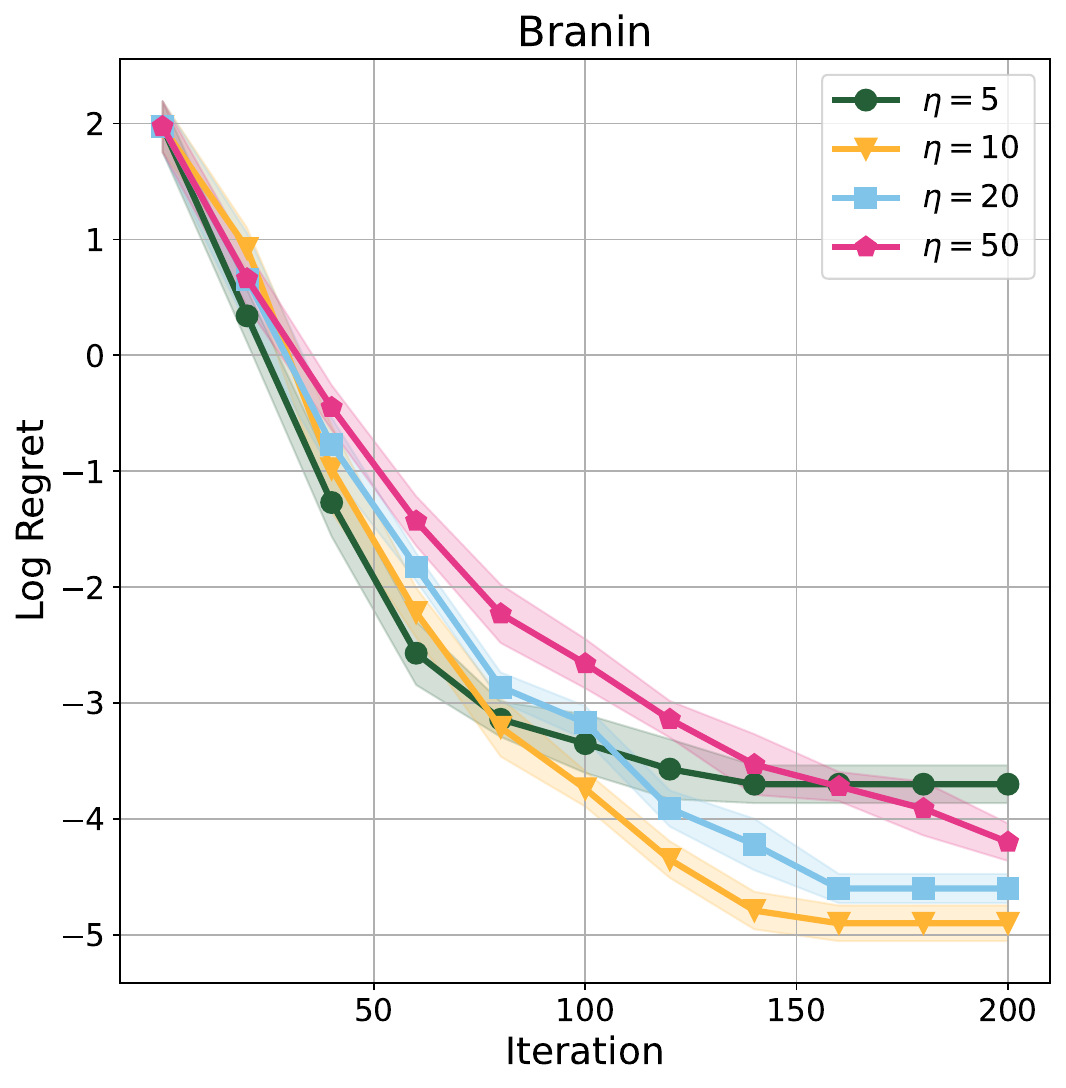}
\includegraphics[width =0.32\textwidth]{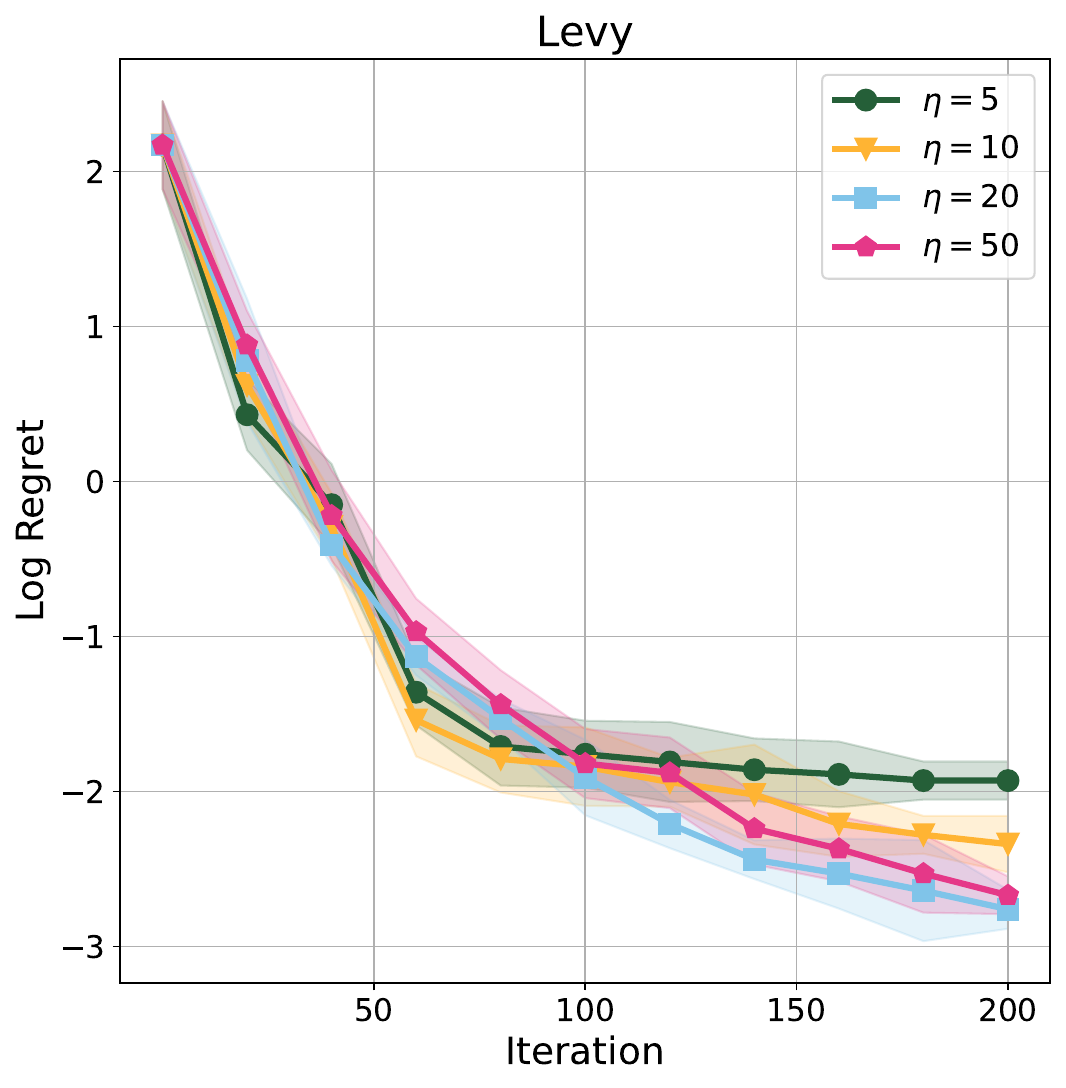}
\includegraphics[width =0.32\textwidth]{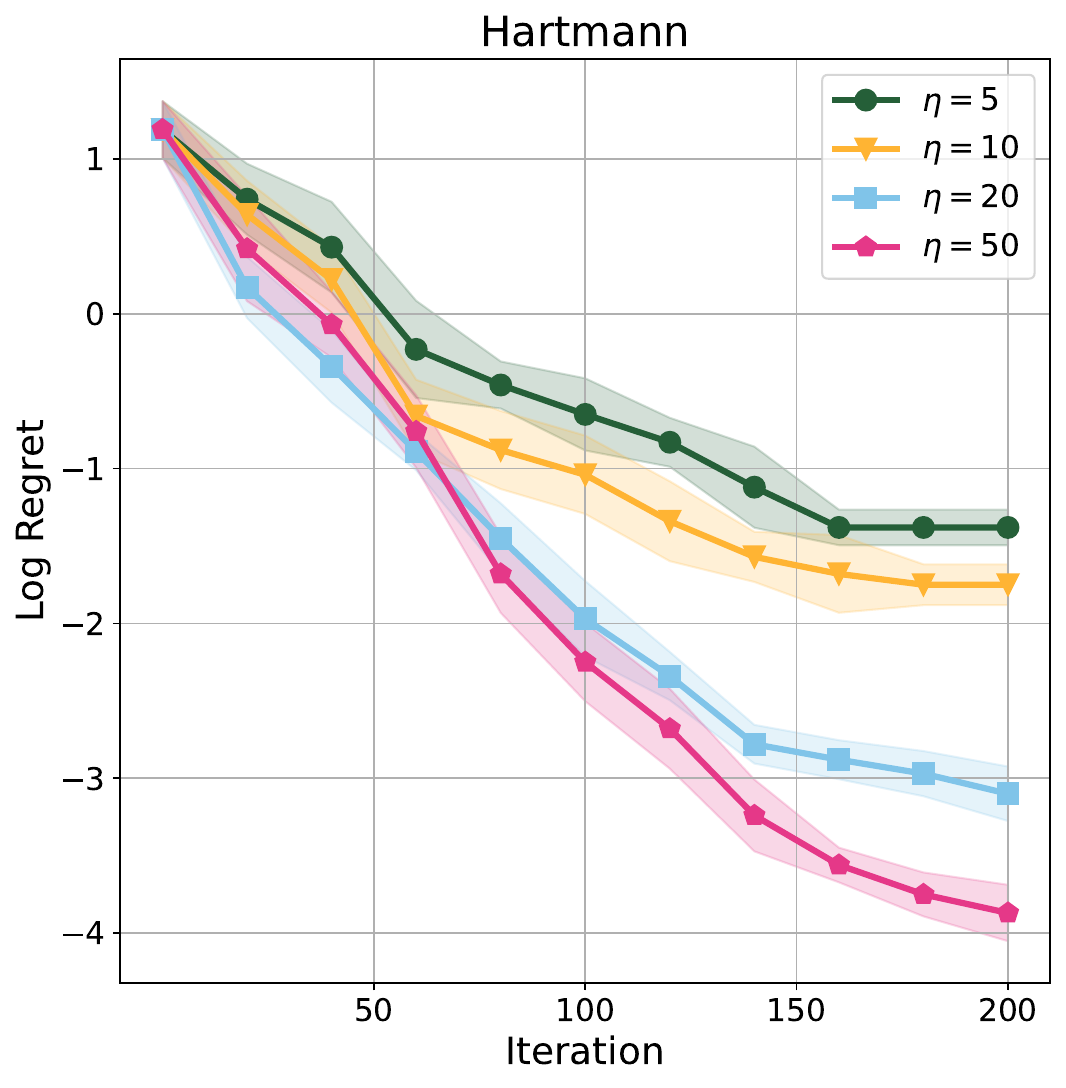}
\caption{Average log regret of FigBO with varying hyperparameter $\eta$ on three different synthetic functions over 20 repetitions. The average log regret and its standard errors are displayed for all tasks.}
\label{fig:hyper}
\end{figure}

The hyperparameter $\eta$ controls the decay rate of the coefficient $\lambda$ in FigBO. A larger $\eta$ results in a slower decay of $\lambda$, causing the algorithm to prioritize exploration for a longer period in the early stages and delaying the transition to the original form of the acquisition function. This extended exploration can be beneficial in complex tasks but may slow convergence if $\eta$ is too large. To evaluate the effect of $\eta$, we vary the value of $\eta$ over $\{5, 10, 20, 50\}$ on three synthetic test functions. The evaluation results are shown in Fig. \ref{fig:hyper}, from which we can observe that as the dimensionality increases, larger values of $\eta$ lead to better model performance and faster convergence. We believe this is because in low-dimensional tasks (e.g., the 2-dimensional Branin function), the model can more easily identify promising search regions, thus requiring less focus on exploration. As a result, smaller values of $\eta$ are sufficient to achieve faster convergence and good model performance. However, in high-dimensional tasks (e.g., the 4-dimensional Levy and the 6-dimensional Hartmann functions), the search space becomes more complex, requiring stronger exploration capabilities to identify promising regions. In such cases, larger values of $\eta$ play a crucial role in enhancing both optimization outcomes and convergence efficiency.

\section{Related Work}
Bayesian optimization has emerged as a powerful framework for optimizing expensive black-box functions, with the choice of acquisition function playing a pivotal role in determining optimization performance. Acquisition functions can be broadly categorized into myopic and non-myopic approaches, depending on whether they consider the future impact of candidate points.

\textbf{Myopic Acquisition Functions.} Myopic acquisition functions focus solely on the immediate improvement expected from evaluating a candidate point, making them computationally efficient and easy to implement. The most widely used myopic acquisition function is expected improvement (EI) \cite{jones1998efficient}, which balances exploration and exploitation by selecting points that maximize the expected improvement over the current best observation. Other popular myopic acquisition functions include upper confidence bound (UCB) \cite{srinivas2012information}, which selects points based on a weighted sum of the posterior mean and standard deviation, with the weight controlling the exploration-exploitation trade-off; probability of improvement (PI) \cite{jones2001taxonomy}, which selects points that maximize the probability of improving upon the current best observation; and Thompson sampling (TS) \cite{thompson1933likelihood}, which randomly samples from the posterior distribution and selects the point with the highest sampled value. Despite their simplicity and effectiveness, myopic acquisition functions lack the ability to consider the long-term impact of candidate points, which can limit their performance in complex optimization tasks \cite{hennig2012entropy, hernandez2014predictive}. To address this limitation, FigBO introduces a look-ahead capability by incorporating the future impact of candidate points on global information gain, enabling myopic acquisition functions to balance immediate improvements with long-term optimization benefits.

\textbf{Non-myopic Acquisition Functions.} Non-myopic acquisition functions aim to address the limitations of myopic approaches by explicitly considering the future impact of candidate points on the optimization process. These methods often involve more complex computations but can yield significant performance improvements in certain settings. Notable non-myopic acquisition functions include entropy search (ES) \cite{hennig2012entropy}, which selects points that maximize the reduction in entropy of the posterior distribution over the global optimum; predictive entropy search (PES) \cite{hernandez2014predictive}, an approximation of ES that reduces computational complexity while maintaining strong performance; max-value entropy search (MES) \cite{wang2017max}, which focuses on maximizing the information gain about the maximum value of the objective function rather than its location; Joint Entropy Search (JES) \cite{hvarfner2022joint} consider the entropy over both the optimum and the noiseless optimal value to improve model performance; and knowledge gradient (KG) \cite{frazier2008knowledge}, which selects points that maximize the expected improvement in the posterior mean at the next iteration. While non-myopic acquisition functions offer theoretical advantages, their practical applicability is often limited by high computational costs and implementation complexity. To address these challenges, FigBO provides a plug-and-play framework that seamlessly integrates with existing myopic acquisition functions, achieving a balance between simplicity and performance. This makes FigBO both practical and effective for complex optimization tasks.

\section{Conclusion and Discussion}
In this paper, we introduced FigBO, a generalized acquisition function framework that enhances myopic acquisition functions with look-ahead capability by incorporating the future impact of candidate points on global information gain. FigBO addresses the limitations of both myopic and non-myopic approaches, offering a practical and effective solution for optimizing expensive black-box functions. Theoretical analysis shows that FigBO retains the convergence properties of standard myopic acquisition functions, while empirical results across synthetic benchmarks and real-world applications demonstrate its superior performance over state-of-the-art methods.

Despite its strengths, FigBO has some limitations. First, our current implementation is based on GP as the surrogate model, and we have not explored its applicability with other surrogate models, such as random forests or Bayesian neural networks. Extending FigBO to support a broader range of surrogate models could further enhance its versatility. Second, FigBO is designed specifically for myopic acquisition functions, and while it provides significant improvements over traditional myopic methods, it does not directly address the challenges of non-myopic acquisition functions. Future work could explore similar design principles to develop new acquisition functions based on non-myopic strategies, potentially combining the strengths of both approaches.

\section*{Statements and Declarations}

\textbf{Author contribution} All authors contributed to the study conception and design. Hui Chen provided the initial idea and wrote the first draft of the manuscript, and also conducted the experiment execution and analysis. Zhangkai Wu assisted with the experiment execution and results analysis. Xuhui Fan and Longbing Cao reviewed and edited the manuscript. All authors read and approved the final manuscript.

\textbf{Conflict of interest} The authors declare that they have no conflict of interest.

\textbf{Data availability} All of them are available in public.

\textbf{Code availability} Code will be released upon acceptance of the paper.

\textbf{Ethics approval} Not applicable.

\textbf{Consent to participate} Not applicable.

\textbf{Consent to publication} Not applicable.

\begin{appendices}

\section{Proof of Theorem 1}

\begin{proof}
To analyze the relationship between the loss functions of $EI_{\Gamma,n}$ and $EI_n$, we first restate Lemma 7 and Lemma 8 from \cite{bull2011convergence}.

\begin{lemma}
For any $m\in \mathbb{N}$, point sequence $\{\mathbf{x}\}^n_{i=1}$, we have following inequality holds for at most $m$ distinct $n$:
\begin{equation}
s_n(\mathbf{x}_{n+1};\ell) \ge Cm^{-(\nu \wedge 1)/d}(\log m)^\beta,\ \ \ \beta:= \begin{cases}\alpha, & \nu \leq 1, \\ 0, & \nu>1,\end{cases}
\end{equation}
where $C>0$ is a constant.
\end{lemma}

\begin{lemma}
Let $\|f\|_{\mathcal{H}_{\ell}(\mathcal{X})} \leq R$, $I_n=(f^*_n-f^*)^+$, $\tau(z)=z\Phi(z)+\phi(z)$ and $s=s_n(\mathbf{x}_{n};\ell)$. Then $EI_n(\mathbf{x})$ is bounded as
\begin{equation}
\label{eq:bound}
\mathrm{max}\bigg(I_n-Rs,\frac{\tau(-R/\sigma)}{\tau(R/\sigma)I_n}\bigg) \leq EI_n(\mathbf{x}) \leq I_n+(R+\sigma)s.
\end{equation}
\end{lemma}

Based on \eqref{eq:bound}, we can trivially determine the both the upper and lower bounds of $EI_{\Gamma,n}$ by incorporating the global uncertainty term $\lambda \Gamma(\mathbf{x})$. The upper bound can be given with $\underset{x\in \mathcal{X}}{\mathrm{max}} \lambda \Gamma(\mathbf{x})$ as
\begin{equation}
\label{eq:upper bound}
EI_{\Gamma,n}-\underset{x\in \mathcal{X}}{\mathrm{max}} \lambda \Gamma(\mathbf{x})=EI_{n}+\lambda \Gamma(\mathbf{x})-\underset{x\in \mathcal{X}}{\mathrm{max}} \lambda \Gamma(\mathbf{x}) \leq EI_n(\mathbf{x}) \leq I_n+(R+\sigma)s.
\end{equation}
Similarly, the lower bound can be directly derived using the $\underset{x\in \mathcal{X}}{\mathrm{min}} \lambda \Gamma(\mathbf{x})$ as 
\begin{equation}
\label{eq:lower bound}
\mathrm{max}\bigg(I_n-Rs,\frac{\tau(-R/\sigma)}{\tau(R/\sigma)I_n}\bigg) \leq EI_n(\mathbf{x}) \leq EI_{\Gamma,n}- \underset{x\in \mathcal{X}}{\mathrm{min}} \lambda \Gamma(\mathbf{x})= EI_{n}+\lambda \Gamma(\mathbf{x})-\underset{x\in \mathcal{X}}{\mathrm{min}} \lambda \Gamma(\mathbf{x}).
\end{equation}
Combining \eqref{eq:upper bound} and \eqref{eq:lower bound}, the $EI_{\Gamma,n}$ is bounded by
\begin{equation}
\mathrm{max}\bigg(I_n-Rs,\frac{\tau(-R/\sigma)}{\tau(R/\sigma)I_n}\bigg) + \underset{x\in \mathcal{X}}{\mathrm{min}} \lambda \Gamma(\mathbf{x}) \leq \mathrm{EI}_{\Gamma,n} (\mathbf{x}) \leq I_n+(R+\sigma)s+\underset{x\in \mathcal{X}}{\mathrm{max}} \lambda \Gamma(\mathbf{x}).
\end{equation}

Furthermore, by the proof of Theorem 2 from \cite{bull2011convergence}, it shows that
\begin{equation}
\sum_n f^*_n-f^*_{n+1} \leq f^*_1-\min f \leq 2\|f\|_{\infty} \leq 2R,
\end{equation}
then for the most $m$ times, we have
\begin{equation}
f^*_n-f^*_{n+1} > 2Rm^{-1}.
\end{equation}
Therefore, we have $s_{n_m}(\mathbf{x}_{n_m+1};\ell) \leq Cm^{-(\nu \wedge 1)/d}(\log m)^\beta$ and $f_{n_m}^*-f(\mathbf{x}_{n_m+1}) \leq 2Rk^{-1}$, where $m \leq n_m \leq 3m$. We can then obtain the bounds on the $EI_{\Gamma,n}$ when $3m \leq m \leq 3(m+1)$ as
\begin{equation}
\begin{aligned}
\mathcal{L}_n(\mathrm{EI}_{\Gamma,n},\mathcal{D}_n,\mathcal{H}_{\ell}(\mathcal{X}),R)&=f_n^*-\min f\\
&\leq f_{n_m}^*-\min f\\
&\leq \frac{\tau(R/\sigma)}{\tau(-R/\sigma)} \big(\mathrm{EI}_{n_m}+\lambda \Gamma(\mathbf{x^*})\big) \\
&= \frac{\tau(R/\sigma)}{\tau(-R/\sigma)} \mathrm{EI}_{\Gamma,n_m}(\mathbf{x}^*)\\
& \leq \frac{\tau(R/\sigma)}{\tau(-R/\sigma)} \mathrm{EI}_{\Gamma,n_m}(\mathbf{x}_{n+1}^*)\\
& \leq \frac{\tau(R/\sigma)}{\tau(-R/\sigma)} \bigg(I_{n_m}+(R+\sigma)s_{n_m}+\underset{x\in \mathcal{X}}{\max} \lambda \Gamma(\mathbf{x})\bigg)\\
& \leq \frac{\tau(R/\sigma)}{\tau(-R/\sigma)} \big(2Rm^{-1}+(R+\sigma)Cm^{-(\nu \wedge 1)/d}(\log m)^\beta\big)\\
&\ \ \ \ +\frac{\tau(R/\sigma)}{\tau(-R/\sigma)} \underset{x\in \mathcal{X}}{\max} \lambda \Gamma(\mathbf{x}).
\end{aligned}
\end{equation}
Here, we represent the second term in the last inequality with $C'=\frac{\tau(R/\sigma)}{\tau(-R/\sigma)} \underset{x\in \mathcal{X}}{\max} \lambda \Gamma(\mathbf{x})$, which indicates the size of the bound beyond $\mathcal{L}_n(\mathrm{EI}_n,\mathcal{D}_n,\mathcal{H}_{\ell}(\mathcal{X}), R)$.
\end{proof}

\section{Proof of Corollary 1}
\begin{proof}
As the number of iterations $n$ approaches infinity, the value of our coefficient $\lambda$ approaches zero. To this end, the ratio between the $\mathcal{L}_n(\mathrm{EI}_{n},\mathcal{D}_n,\mathcal{H}_{\ell}(\mathcal{X}),R)$ and $\mathcal{L}_n(\mathrm{EI}_{\Gamma,n},\mathcal{D}_n,\mathcal{H}_{\ell}(\mathcal{X}),R)$ can be expressed as
\begin{equation}
\begin{aligned}
&\underset{n \rightarrow \infty}{\lim} \frac{\mathcal{L}_n(\mathrm{EI}_{\Gamma,n},\mathcal{D}_n,\mathcal{H}_{\ell}(\mathcal{X}),R)}{\mathcal{L}_n(\mathrm{EI}_{n},\mathcal{D}_n,\mathcal{H}_{\ell}(\mathcal{X}),R)}\\
\leq & 1+ \underset{n \rightarrow \infty}{\lim} \frac{\frac{\tau(R/\sigma)}{\tau(-R/\sigma)} \underset{x\in \mathcal{X}}{\mathrm{max}} \lambda \Gamma(\mathbf{x})}{\frac{\tau(R/\sigma)}{\tau(-R/\sigma)} \big(2Rm^{-1}+(R+\sigma)Cm^{-(\nu \wedge 1)/d}(\log m)^\beta\big)+\frac{\tau(R/\sigma)}{\tau(-R/\sigma)} \underset{x\in \mathcal{X}}{\mathrm{max}} \lambda \Gamma(\mathbf{x})}\\
=& 1+ \underset{n \rightarrow \infty}{\lim} \frac{\frac{\tau(R/\sigma)}{\tau(-R/\sigma)} \underset{x\in \mathcal{X}}{\mathrm{max}} \frac{\eta}{n} \Gamma(\mathbf{x})}{\frac{\tau(R/\sigma)}{\tau(-R/\sigma)} \big(2Rm^{-1}+(R+\sigma)Cm^{-(\nu \wedge 1)/d}(\log m)^\beta\big)+\frac{\tau(R/\sigma)}{\tau(-R/\sigma)} \underset{x\in \mathcal{X}}{\mathrm{max}} \lambda \Gamma(\mathbf{x})}\\
=&1.
\end{aligned}
\end{equation}
\end{proof}

\section{Experimental Details}
\subsection{GP Prior Samples}
For generating the GP sample tasks, we employed the random Fourier features method \cite{rahimi2007random}, where the weights are sampled from the spectral density corresponding to a squared exponential kernel. Each task is defined by specific hyperparameters: the length scale $\ell$ (which varies across dimensions for a fair comparison), the signal variance $\sigma_f^2$, and the noise variance $\sigma_{\epsilon}^2$. These values are detailed in Table \ref{tab:hyper_prior}, along with an approximate range of the output values for each GP sample. Since the true optimal values for these tasks are unknown, we approximate them using a combination of extensive random search and subsequent local optimization on the most promising candidate points \cite{hvarfner2022joint}.

\begin{table}[t]
\renewcommand{\arraystretch}{1.0}
\caption{Hyperparameter setting for GP prior samples tasks.}
\label{tab:hyper_prior}
\centering
% \resizebox{\linewidth}{!}{
\begin{tabular}{ccccc}
\toprule
 $D$ & $\ell$ & $\sigma_f^2$ & $\sigma_{\epsilon}^2$ & Approximate Range\\
 \midrule
2& 0.1 & 10 & 0.01 & [-9, 9] \\ 
4 & 0.2 & 10 & 0.01 & [-11, 11]\\
6 & 0.3 & 10 & 0.01 & [-13, 13]\\
12 & 0.6 & 10 & 0.01 & [-18, 18]\\
\bottomrule
\end{tabular}
\end{table}

\subsection{Synthetic Test Functions}
We choose three synthetic acquisition functions with different dimensions to evaluate the model performance.  Specifically, we consider the Branin function, a two-dimensional benchmark function with three global optimum; the Levy function, a four-dimensional function with one global optimum; and Hartmann function, a six-dimensional function with one global optimum. Table \ref{tab:hyper_synthetic} provides more details about these benchmark functions.

\subsection{MLP Classification Tasks}
The MLP classification tasks involve optimizing four hyperparameters for a multi-layer perceptron model: L2 regularization (alpha), batch size, initial learning rate, and network width. The number of layers (depth) is fixed at $2$ because its limited range of integer values makes optimization less meaningful, while the other two integer-valued hyperparameters—batch size and width—have significantly larger domains, allowing them to be treated as continuous during optimization. Further details about the hyperparameters are provided in Table \ref{tab:hyper_MLP}. All hyperparameters are scaled to the $[0, 1]$ range, transformed, and rounded to the nearest integer when necessary during objective function evaluation. Additionally, non-fixed parameters are evaluated in log scale.

\begin{table}[t]
\renewcommand{\arraystretch}{1.0}
\caption{Benchmark used for synthetic tasks.}
\label{tab:hyper_synthetic}
\centering
% \resizebox{\linewidth}{!}{
\begin{tabular}{cccc}
\toprule
 Task & Dimension & $\sigma_{\epsilon}^2$ & Search Space\\
 \midrule
Branin & 2 & 0.01 & [-5, 10] $\times$ [0, 15] \\
Levy & 4 & 0.01 & [-10, 5] $\times$ [-10, 10] $\times$ [-5, 10] $\times$ [-1, 10] \\
Hartmann & 6 & 0.01 & $[0, 1]^D$ \\
\bottomrule
\end{tabular}
\end{table}

\begin{table}[t]
\renewcommand{\arraystretch}{1.0}
\caption{Hyperparameter used for MLP tasks.}
\label{tab:hyper_MLP}
\centering
% \resizebox{\linewidth}{!}{
\begin{tabular}{ccc}
\toprule
 Task & Type & Search Range\\
 \midrule
Alpha (L2) & Continuous & $[10^{-8}, 10^{-3}]$ \\
Batch Size & Integer & $[2^2, 2^8]$\\
Depth & Fixed to 2 & $\{1, 2, 3\}$ \\
Initial Learning Rate & Continuous & $[10^{-5}, 1]$ \\
Width & Integer & $[2^4, 2^{10}]$ \\
\bottomrule
\end{tabular}
\end{table}

\end{appendices}


\bibliography{sn-bibliography}% common bib file

\end{document}